\documentclass[runningheads]{llncs}
\usepackage{graphicx}
\usepackage{comment}
\usepackage{amsmath,amssymb} 
\usepackage{color}
\usepackage{bbding}
\usepackage{pifont}
\usepackage{xcolor}
\usepackage{colortbl}
\usepackage{mathtools}
\usepackage{mathrsfs}
\usepackage{multirow}
\usepackage{algorithmic}
\usepackage{subfigure}
\usepackage{array}
\usepackage{tabu}
\usepackage{adjustbox}
\usepackage{tabularx}
\usepackage{booktabs}
\usepackage{diagbox}
\usepackage{array}
\usepackage{float}
\usepackage{threeparttable}
\usepackage{authblk}
\usepackage{makecell}
\usepackage[width=122mm,left=12mm,paperwidth=146mm,height=193mm,top=12mm,paperheight=217mm]{geometry}
\usepackage[pagebackref=true,breaklinks=true,letterpaper=true,colorlinks,bookmarks=false,urlcolor=blue]{hyperref}
\newcommand{\textBC}[2]{\textbf{\textcolor{#1}{#2}}}
\begin{document}
	\pagestyle{headings}
	\mainmatter
	\def\ECCVSubNumber{2852}  
	
	\title{Suppress and Balance: A Simple Gated Network for Salient Object Detection} 

	\titlerunning{Suppress and Balance}
	%
	\author{Xiaoqi Zhao\inst{1}\protect\footnotemark[4], Youwei Pang\inst{1}\protect\footnotemark[4], Lihe Zhang\inst{1}\thanks{Corresponding author.}, Huchuan Lu\inst{1,2}, and Lei Zhang\inst{3,4}}
	\authorrunning{Zhao et al.}
	%
	\institute{Dalian University of Technology, China 
		\and
		Peng Cheng Laboratory 
		\and 
		Dept. of Computing, The Hong Kong Polytechnic University, China
		\and 
		DAMO Academy, Alibaba Group \\
		\email{\{zxq,lartpang\}@mail.dlut.edu.cn, \{zhanglihe,lhchuan\}@dlut.edu.cn,\ cslzhang@comppolyu.edu.hk}
		\url{https://github.com/Xiaoqi-Zhao-DLUT/GateNet-RGB-Saliency}
	}
	\maketitle
	\renewcommand{\thefootnote}{\fnsymbol{footnote}} 
	\footnotetext[4]{These authors contributed equally to this work.} 
	\renewcommand{\thefootnote}{\arabic{footnote}}
	\begin{abstract}
		Most salient object detection approaches use U-Net or feature pyramid networks (FPN) as their basic structures. These methods ignore two key problems when the encoder exchanges information with the decoder: one is the lack of interference control between them, the other is without considering the disparity of the contributions of different encoder blocks. In this work, we propose a simple gated network (GateNet) to solve both issues at once. With the help of multilevel gate units, the valuable context information from the encoder can be optimally transmitted to the decoder. We design a novel gated dual branch structure to build the cooperation among different levels of features and improve the discriminability of the whole network. Through the dual branch design, more details of the saliency map can be further restored. In addition, we adopt the atrous spatial pyramid pooling based on the proposed ``Fold'' operation (Fold-ASPP) to accurately localize salient objects of various scales. Extensive experiments on five challenging datasets demonstrate that the proposed model performs favorably against most state-of-the-art methods under different evaluation metrics.
	\end{abstract}

	\section{Introduction}
	Salient object detection aims to identify the visually distinctive regions or objects in a scene and then accurately segment them. In many computer vision applications, it is used as a pre-processing step, such as scene classification~\cite{classification}, visual tracking~\cite{tracking}, person re-identification~\cite{Reid}, light field image segmentation~\cite{LFSD_CNNs} and image captioning~\cite{Imagecaption}, etc. 
	\begin{figure}
		\centering
		\includegraphics[width=0.55\linewidth]{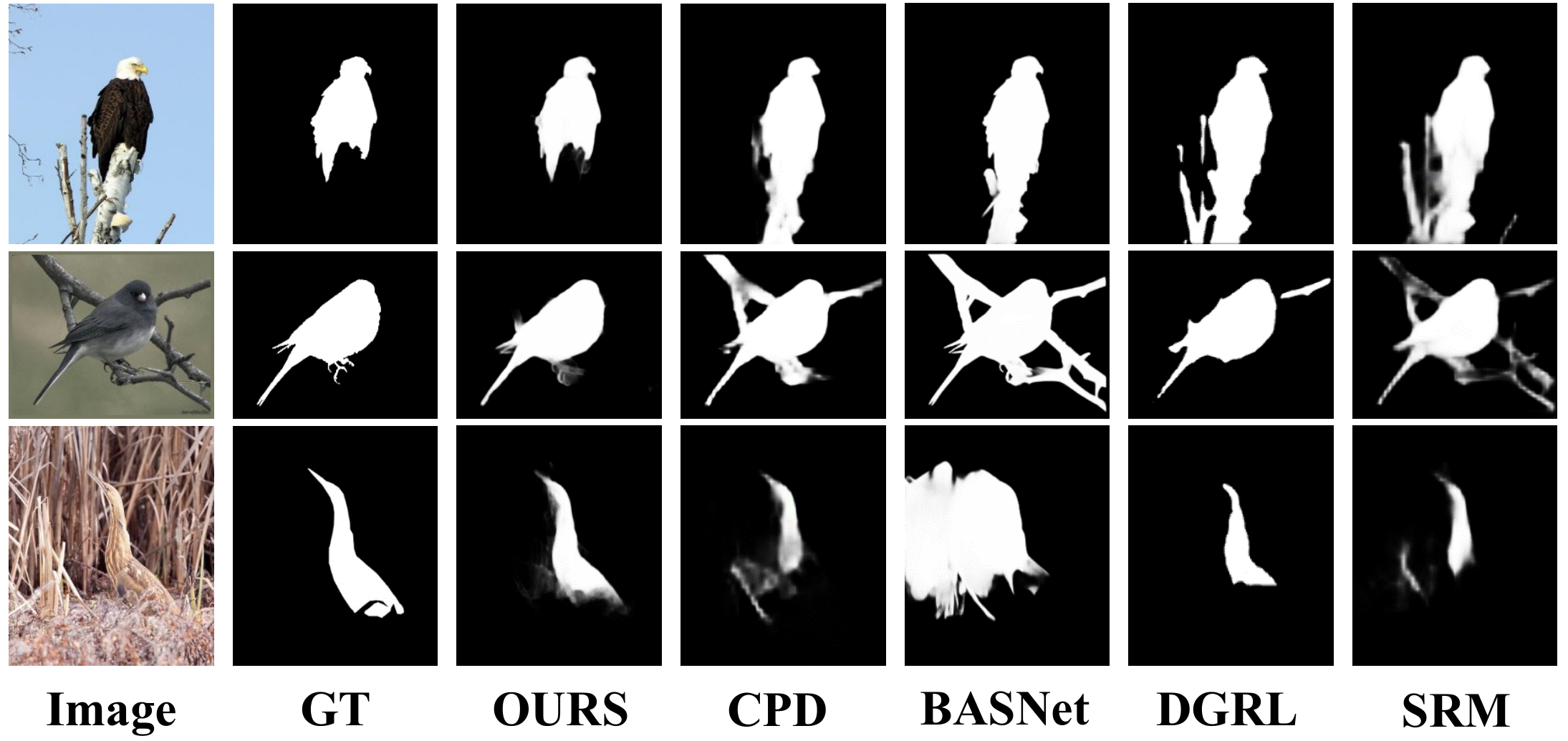}
		\caption{Visual comparison of different CNN based methods.}\label{fig:Qualitative1}
	\end{figure}
	
	With the development of deep learning, salient object detection has gradually evolved from the traditional method based on manual design features to the deep learning method. In recent years, U-shape based structures~\cite{Unet,FPN} have received the most attention due to their ability to utilize multilevel information to reconstruct high-resolution feature maps. Therefore, most state-of-the-art saliency detection networks~\cite{DHS,DSS,Amulet,DGRL,BMPM,PAGRN,PASE,BASNet,MINet} adopt U-shape as the encoder-decoder architecture. And many methods aim at combining multilevel features in either the encoder~\cite{Amulet,DGRL,BMPM,PASE,BASNet,CPD} or the decoder~\cite{DHS,DSS,PAGRN,CPD}. For each convolutional block, they separately formulate the relationships of internal features for forward update. It is well known that the high-quality saliency maps predicted in the decoder rely heavily on the effective features provided by the encoder. Nevertheless, the aforementioned methods directly use an all-pass skip-layer structure to concatenate the features of the encoder to the decoder, and the effectiveness of feature aggregation at different levels is not quantified. These restrictions not only introduce misleading context information into the decoder but also result in that the really useful features can not be adequately utilized. In cognitive science, Yang \textit{et al.}~\cite{NM} show that inhibitory neurons play an important role in how the human brain chooses to process the most important information from all the information presented to us. And inhibitory neurons ensure that humans respond appropriately to external stimuli by inhibiting other neurons and balancing excitatory neurons that stimulate neuronal activity. Inspired by this work, we think that it is necessary to set up an information screening unit between each pair of encoder and decoder blocks in saliency detection. It can help distinguish the most intense features of salient regions and suppress background interference, as shown in Fig.~\ref{fig:Qualitative1}, in which these images have easily-confused backgrounds or low-contrast objects.
	
	Moreover, due to the limited receptive field, a single-scale convolutional kernel is difficult to capture context information of size-varying objects. This motivates some efforts~\cite{R3Net,BMPM} to investigate multiscale feature extraction. These methods directly equip an atrous spatial pyramid pooling module~\cite{Deeplab} (ASPP) in their networks. However, when using a convolution with a large dilation rate, the information under the kernel seriously lacks correlation due to inserting too many zeros. This may be detrimental to the discrimination of subtle image structures.
	
	In this paper, we propose a simple gated network (GateNet) for salient object detection. Based on the feature pyramid network (FPN), we construct multilevel gate units to combine the features from the decoder and the encoder. We use convolution operation and nonlinear functions to calculate the correlations among features and assign gate values to different blocks. In this process, a partnership is established between different blocks by using weight distribution and the decoder can obtain more efficient information from the encoder and pay more attention to the salient regions. Since the top-layer features of the encoder network contain rich contextual information, we construct a folded atrous spatial pyramid pooling (Fold-ASPP) module to gather multiscale high-level saliency cues. With the ``Fold'' operation, the atrous convolution is implemented on a group of local neighborhoods rather than a group of isolated sampling points, which can help generate more stable features and more adequately depict finer structure. In addition, we design a parallel branch by concatenating the output of the FPN branch and the features of the gated encoder, so that the residual information complementary to the FPN branch is supplemented to generate the final saliency map. 
	
	Our main contributions can be summarized as follows.
	\begin{itemize}
		\item We propose a simple gated network to adaptively control the amount of information that flows into the decoder from each encoder block. With multilevel gate units, the network can balance the contribution of each encoder block to the the decoder block and suppress the features of non-salient regions.
		
		\item We design a Fold-ASPP module to capture richer context information and localize salient objects of various sizes. By the ``Fold'' operation, we can obtain more effective feature representation.
		
		\item We build a dual branch architecture. They form a residual structure, complement each other through the gated processing and generate better results.    
	\end{itemize}
	
	We compare the proposed model with seventeen state-of-the-art methods on five challenging datasets. The results show that our method performs much better than other competitors. And, it achieves a real-time speed of 30 fps.
	
	\section{Related Work}
	\subsection{Salient Object Detection}
	Early saliency detection methods are based on low-level features and some heuristics prior knowledge, such as color contrast~\cite{colorcontrast/Fm}, background prior~\cite{boundarybackground} and center prior~\cite{centerprior}. Most of them using hand-crafted features, and more details about the traditional methods are discussed in~\cite{Survey}.
	
	With the breakthrough of deep learning in the field of computer vision, a large number of convolutional neural networks-based salient object detection methods have been proposed and their performance had been improved gradually. 
	Especially, fully convolutional networks (FCN), which avoid the problems caused by the fully-connected layer, become the mainstream for dense prediction tasks. Wang \textit{et al.}~\cite{RFCN} use weight sharing methods to iteratively refine features and promote mutual fusion between features. Hou \textit{et al.}~\cite{DSS} achieve efficient feature expression by continuously blending features from deep layers into shallow layers. However, the single-scale feature cannot roundly characterize various objects as well as image contexts. How to get multiscale features and integrate context information is an important problem in saliency detection.
	\subsection{Multiscale Feature Extraction}
	Recently, the atrous spatial pyramid pooling module (ASPP)~\cite{Deeplab} is widely applied in many tasks and networks. The atrous convolution can enlarge the receptive field to obtain large-scale features and does not increase the computational cost. 
	Therefore, it is often used in saliency detection networks. 
	Zhang \textit{et al.}~\cite{BMPM} insert several ASPP modules into the encoder blocks of different levels, while Deng \textit{et al.}~\cite{R3Net} install it on the highest-level encoder block. 
	Nevertheless, the repeated stride and pooling operations already make the top-layer features lose much fine information. With the increase of atrous rate, the correlation of sampling points further degrades, which leads to difficulties in capturing the changes of image details (e.g., lathy background regions between adjacent objects or spindly parts of objects). In this work, we propose a folded ASPP to alleviate these issues and achieve a \textsl{local-in-local} effect.
	\subsection{Gated Mechanisms}
	The gated mechanism plays an important role in controlling the flow of information and is widely used in the long short term memory (LSTM). 
	In~\cite{gatedlabel}, the gate unit combines two consecutive feature maps of different resolutions from the encoder to generate rich contextual information. Zhang \textit{et al.}~\cite{BMPM} adopt gate function to control the message passing when combining feature maps at all levels of the encoder. 
	Due to the ability to filter information, the gated mechanism can also be seen as a special kind of attention mechanism.
	Some saliency methods~\cite{RAS,PAGRN,PASE} employ attention networks. Zhang \textit{et al.}~\cite{PAGRN} apply both spatial and channel attention to each layer of the decoder. Wang \textit{et al.}~\cite{PASE} exploit the pyramid attention module to enhance saliency representations for each layer in the encoder and enlarge the receptive field. 
	The above methods all unilaterally consider the information interaction between different levels either in the encoder or in the decoder. We integrate the features from the encoder and the decoder to formulate gate function, which plays the role of block-wise attention and model the overall distribution of all blocks in the network from the global perspective. While previous methods actually utilize the block-specific feature to compute dense attention weights for the corresponding block.
	Moreover, in order to take advantage of rich contextual information in the encoder, these methods directly feed the encoder features into the decoder and do not consider their mutual interference. Our proposed gate unit can naturally balance their contributions, thereby suppressing the response of the encoder to non-salient regions. 
	Experimental results in Fig.~\ref{fig:Gate_value} and Fig.~\ref{fig:Gate_suppress} intuitively demonstrate the effect of multilevel gate units on the above two aspects, respectively.
	
	\section{Proposed Method}
	\begin{figure*}
		\includegraphics[width=\textwidth, height=0.5\linewidth]{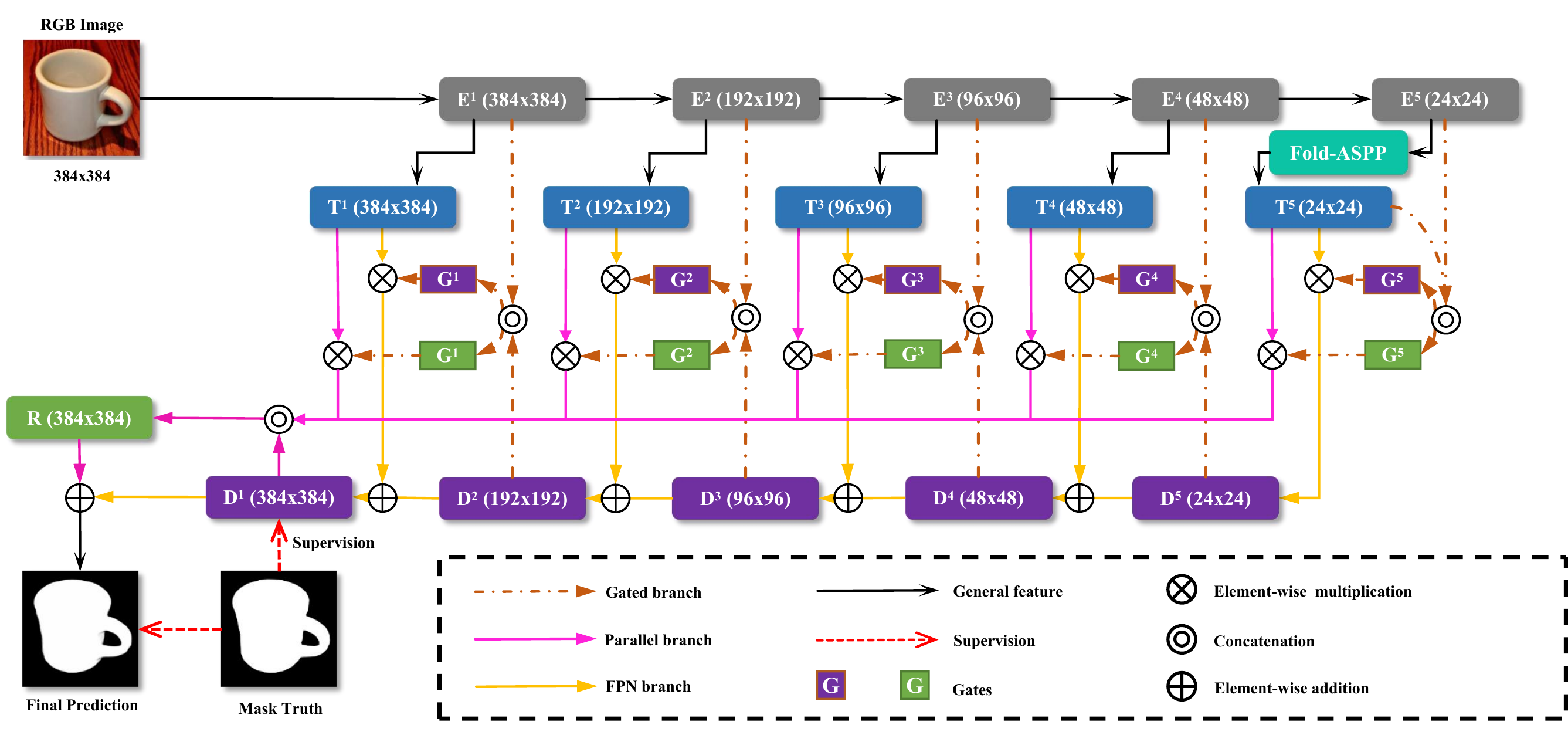}\\ 
		\centering
		\caption{The overall architecture of the gated network. It consists of   the VGG-16 encoder ($\mathbf{E}^1 \sim \mathbf{E}^5$), five transition layers ($\mathbf{T}^1 \sim \mathbf{T}^5$), five gate units ($\mathbf{G}^1 \sim \mathbf{G}^5$), five decoder blocks ($\mathbf{D}^1 \sim \mathbf{D}^5$) and the {Fold-ASPP} module. We employ twice supervision in this network. Once acts at the end of the FPN branch ${D}^1$. The other is used to guide the fusion of the two branches.} 		
		\label{fig:GateNet}
	\end{figure*} 
	The gated network architecture is shown in Fig.~\ref{fig:GateNet}, in which encoder blocks, transition layers, decoder blocks and  gate units are respectively denoted as $\mathbf{E}^i$, $\mathbf{T}^i$ , $\mathbf{D}^i$ and $\mathbf{G}^i$  ($i \in \left \{1, 2, 3, 4, 5 \right \}$ indexes different levels). And their output feature maps are denoted as $E^i$, $T^i$, $D^i$ and $G^i$, respectively. 
	The final prediction is obtained by combining the FPN branch and the parallel branch.  
	In this section, we first describe the overall architecture, then detail the gated dual branch structure and the folded atrous spatial pyramid pooling module.
	\subsection{Network Overview}\label{sec:Network}
	\noindent\textbf{Encoder Network.} In our model, the encoder is based on a common pretrained backbone network, \textit{e.g.}, the VGG~\cite{VGG}, ResNet~\cite{Resnet} or ResNeXt~\cite{ResNext}. We take the VGG-16 network as an example, which contains thirteen Conv layers, five max-pooling layers and two fully connected layers. In order to fit saliency detection task, similar to most previous approaches~\cite{Amulet,DSS,PAGRN,BMPM}, we cast away all the fully-connected layers of the VGG-16 and remove the last pooling layer to retain details of last convolutional layer. 
	
	\noindent\textbf{Decoder Network.} The decoder comprises three main components. i) The FPN branch, which continually fuses different level features from ${T}^1 \sim{T}^5$ by element-wise addition. ii) The parallel branch, which combines the saliency map of the FPN branch with the feature maps of transition layers by cross-channel concatenation. At the same time, multilevel gate units ($\mathbf{G}^1 \sim \mathbf{G}^5$) are inserted between the transition layer and the decoder layer. 
	iii) The Fold-ASPP module, which improves the original atrous spatial pyramid pooling (ASPP) by using a ``Fold'' operation. It can take advantage of semantic features learned from ${E}^5$ to provide multiscale information to the decoder.

	\subsection{Gated Dual Branch}\label{sec:Gated Dual Branch}

	\begin{figure}[t]
		\centering
		\includegraphics[width=0.6\linewidth,height=0.1\linewidth]{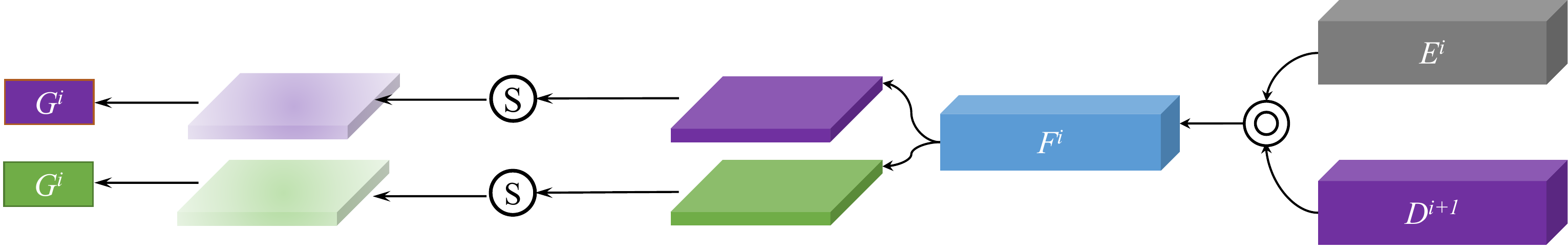}
		\caption{Detailed illustration of the gate unit. ${E}^i$, ${D}^{i+1}$ indicates feature maps of the current encoder block and those of the previous decoder block, respectively. \textcircled{\scriptsize S} is sigmoid function.}\label{fig:Gate_Unit}
	\end{figure}
	
	\begin{figure}[t]
		\centering
		\includegraphics[width=0.6\linewidth]{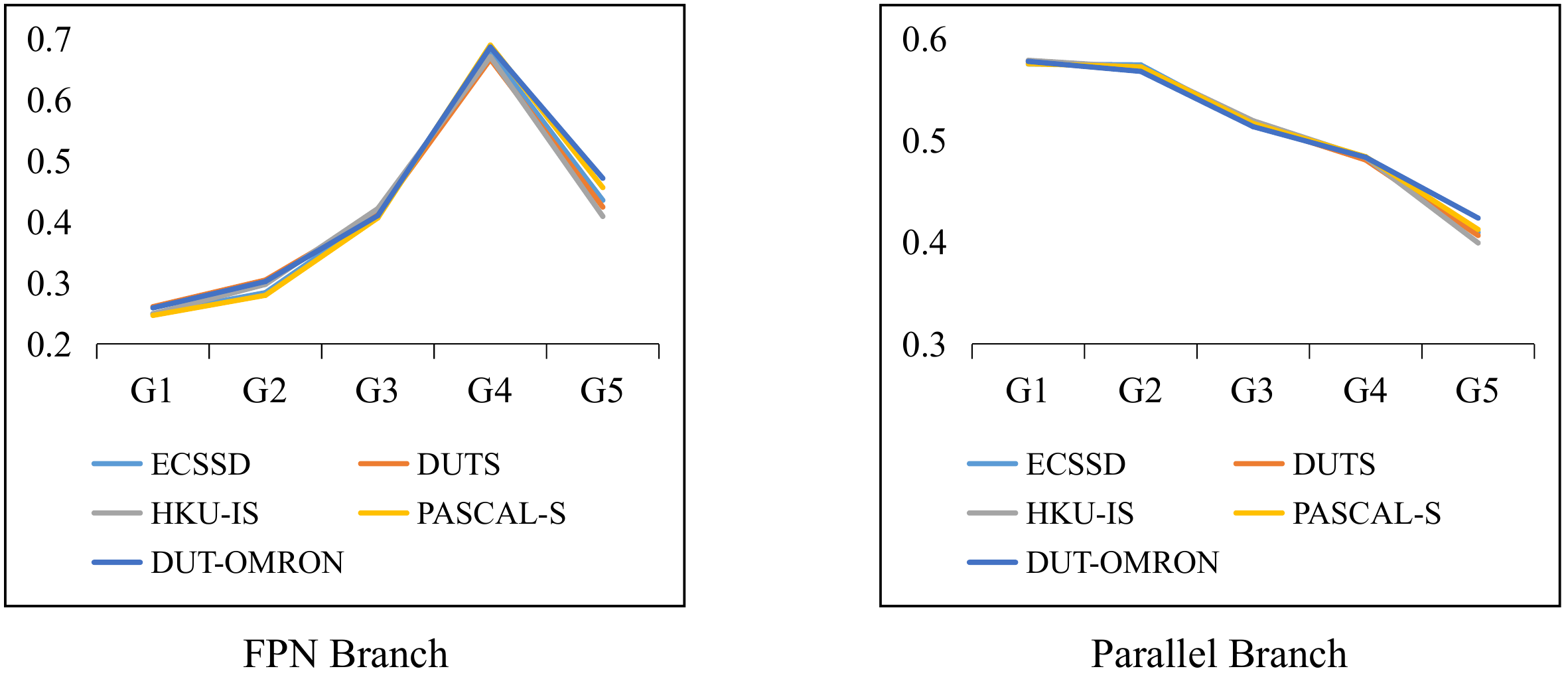}
		\caption{The distributions of the gate weights on  five datasets. We calculate the average gate values for each level of the FPN branch and the parallel branch across all images in every dataset. For the FPN branch, the low-level gate values are significantly smaller than the high-level ones. For the parallel branch, the gate values gradually decrease with the promotion of levels.}\label{fig:Gate_value}
	\end{figure}
	
	The gate unit can control the message passing between scale-matching encoder and decoder blocks. By combining the feature maps of the previous decoder block, the gate value also characterizes the contribution that the current block of the encoder can provide. 
	Fig.~\ref{fig:Gate_Unit} shows the internal structure of the proposed gate unit. In particular, encoder feature ${E}^i$ and decoder feature ${D}^{i+1}$ are integrated to obtain feature ${F}^i$, and then it is fed into two branches, which includes a series of convolution, activation and pooling operations, to compute a pair of gate values ${G}^i$.   
	The entire gated process can be formulated as,
	\begin{equation}\label{equ:3}
	{G}^i
	= \left\{\begin{matrix}
	P(S(Conv(Cat(E^i, D^{i+1})))) & \text{ if } i=1, 2, 3, 4\\
	P(S(Conv(Cat(E^i, T^i)))) & \text{ if } i=5
	\end{matrix}\right.
	\end{equation}
	where $Cat(\cdot)$ is the concatenation operation among channel axis, $Conv(\cdot)$ refers to the convolution layer, $S(\cdot)$ is the element-wise sigmoid function, and $P(\cdot)$ is the global average pooling. The output channel of $Conv(\cdot)$ is 2. The resulted gate vector ${G}^i$ has two different elements which correspond to two gate values in Fig.~\ref{fig:Gate_Unit}.
	
	Given the gate values, they are applied to the FPN branch and the parallel branch for weighting the transition-layer features ${T}^1 \sim {T}^5$, which are generated by exploiting $3 \times 3$ convolution to reduce the dimension of ${E}^1 \sim {E}^4$ and the Fold-ASPP to finely process ${E}^5$ (Please see Fig.~\ref{fig:GateNet} for details). Through multilevel gate units, we can suppress and balance the information flowing from different encoder blocks to the decoder.

	In Fig.~\ref{fig:Gate_value}, we statistically demonstrate the curves of gate value with a convolutional level as the horizontal axis. 
	It can be seen that the high-level encoder features contribute more contextual guidance to the decoder than the low-level encoder features in the FPN branch. This trend is just the opposite in the parallel branch. It is because the FPN branch is responsible to predict the main body of the salient object by progressively combining multilevel features, which needs more high-level semantic knowledge. While the parallel branch, as a residual structure, aims to fill in the details, which are mainly contained in the low-level features. 
	In addition, some visual examples are shown in Fig.~\ref{fig:Gate_suppress} demonstrate that multilevel gate units can significantly suppress the interference from each encoder block and enhance the contrast between salient and non-salient regions. Since the proposed gate unit is simple yet effective, a raw FPN network with multilevel gate units can be viewed as a new baseline for saliency detection task.
	\begin{figure}[t]
		\centering
		\includegraphics[width=0.6\linewidth]{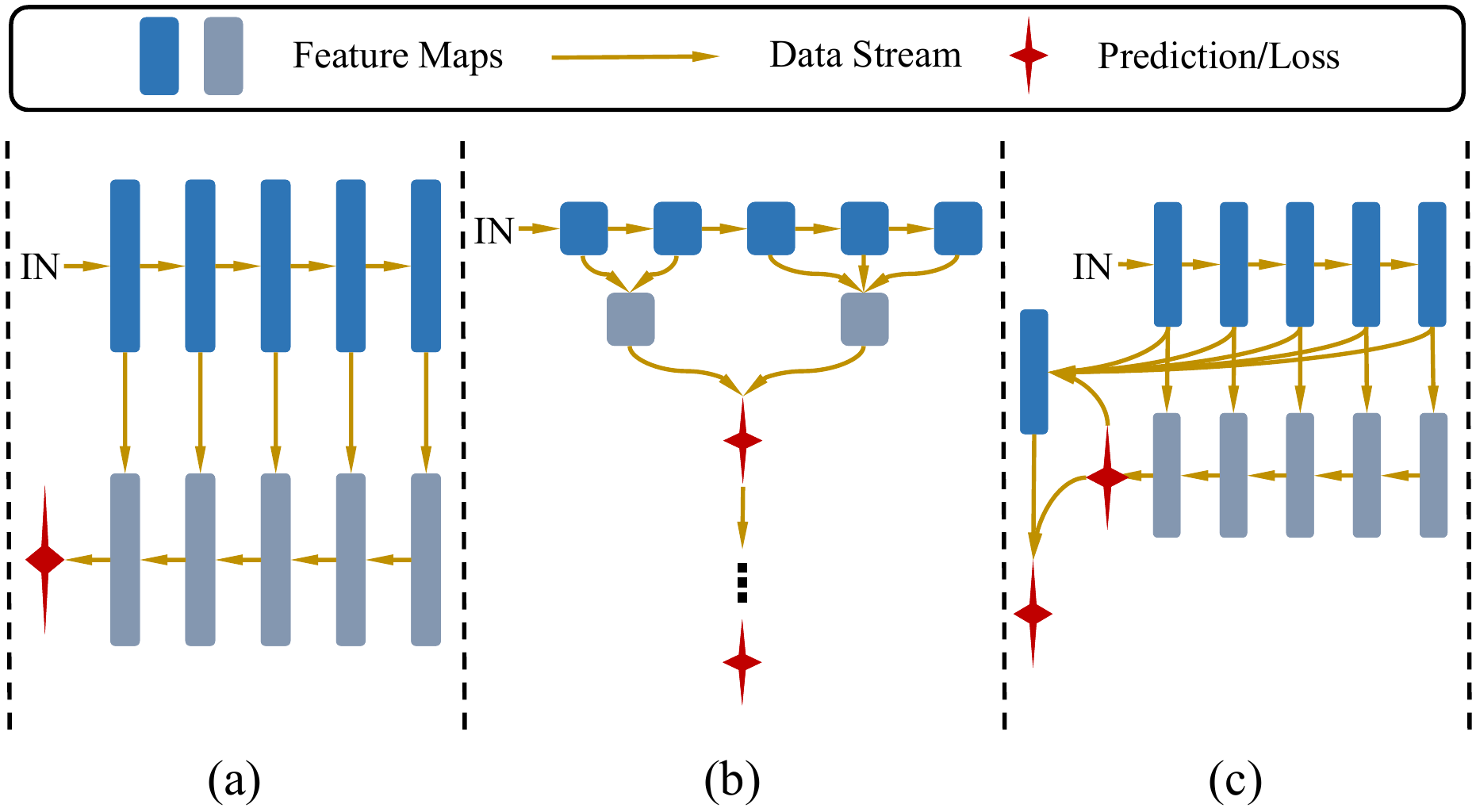}\\
		
		\caption{Illustration of different decoder architectures. (a) Progressive structure, (b) Parallel structure and (c) Our dual branch structure.}
		\label{fig:Dual_branch}
	\end{figure} 
	Most existing models either use progressive decoder~\cite{BMPM,DGRL,PAGRN,PASE} or parallel decoder~\cite{R3Net,PFA}, as shown in Fig.~\ref{fig:Dual_branch}. The progressive structure  begins with the top layer and gradually utilizes the output of the higher layer as prior knowledge to fuse the encoder features. This mechanism is not conducive to the recovery of details because the high-level features lack fine information. While the parallel structure easily results in inaccurate localization of objects since the low-level features without semantic information directly interfere with the capture of global structure cues. In this work, we mix the two structures to build a dual branch decoder to overcome the above restrictions.
	We briefly describe the FPN branch. Taking $D^i$ as an example, we firstly apply bilinear interpolation to upsample the higher-level feature $D^{i+1}$ to the same size as ${T}^{i}$. Next, to decrease the number of parameters, ${T}^{i}$ is reduced to $32$ channels and fed into gate unit $G^{i}$. Lastly, the gated feature is fused with the upsampled feature of $D^{i+1}$ by element-wise addition and convolutional layers. This process can be formulated as follows:
	\begin{equation}\label{equ:fpn}
	D^i = \left\{\begin{matrix}
	Conv(G_1^i \cdot T^i + Up(D^{i+1})) & \text{ if } i=1, 2, 3, 4\\
	Conv(G_1^i \cdot T^i) & \text{ if } i=5,
	\end{matrix}\right.
	\end{equation}
	where $D^{1}$ is a single-channel feature map with the same size as the input image. 
	
	In the parallel branch, we firstly upsample ${T}^1 \sim {T}^5$ to the same size of $D^1$. Next, the multilevel gate units are followed to weight the corresponding transition-layer features. Lastly, we combine $D^1$ and the gated features by cross-channel concatenation.
	The whole process is written as follows:
	\begin{equation}\label{equ:1}
	\begin{split}
	F_{Cat} = Cat(&D^1, Up(G_2^1 \cdot T^1), Up(G_2^2 \cdot T^2),\\
	&  Up(G_2^3 \cdot T^3), Up(G_2^4 \cdot T^4), Up(G_2^5 \cdot T^5)) .
	\end{split} 
	\end{equation}
	
	The final saliency map $S^F$ is generated by integrating the predictions of the two branches with a residual connection as shown in Fig.~\ref{fig:Dual_branch}(c),
	\begin{equation}\label{equ:2}
	\begin{split}
	S^F = S(Conv(F_{Cat}) + D^1)),
	\end{split}
	\end{equation}
	where $S(\cdot)$ is the element-wise sigmoid function.
	\subsection{Folded Atrous Spatial Pyramid Pooling}\label{sec:Fold}
\begin{figure}[t]
		\centering
		\includegraphics[width=0.55\linewidth]{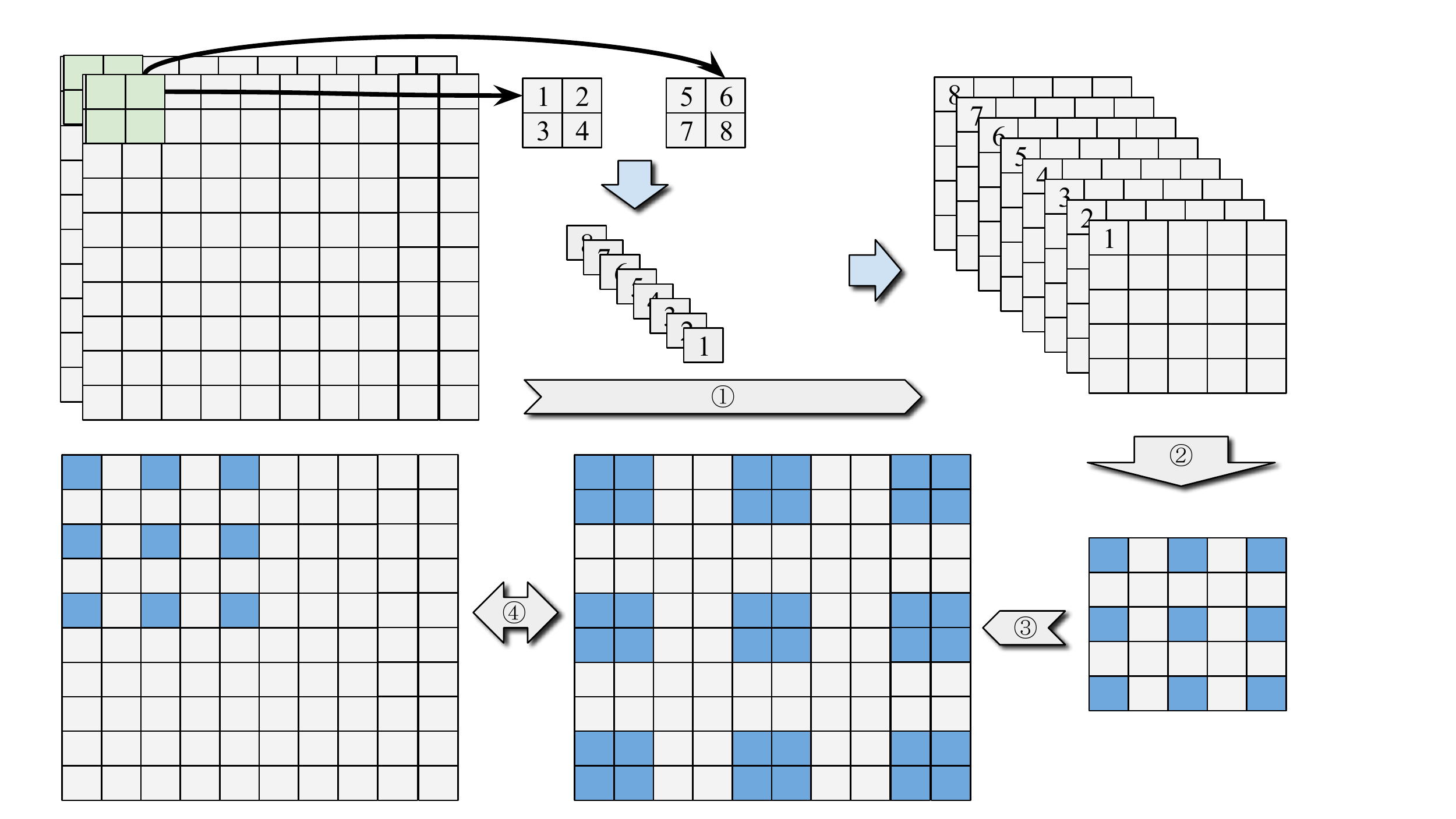}\\
		
		\caption{Illustration of the folded convolution. We use \textcircled{\scriptsize 1}, \textcircled{\scriptsize 2} and \textcircled{\scriptsize 3} to respectively indicate ``Fold'' operation, atrous convolution and ``Unfold'' operation. \textcircled{\scriptsize 4} shows the comparison between atrous convolution (Left) and the folded atrous convolution (Right).}
		\label{fig:FoldConv}
	\end{figure}

	In order to obtain robust segmentation results by integrating multiscale information, atrous spatial pyramid pooling (ASPP) is proposed in Deeplab~\cite{Deeplab}. And some works~\cite{BMPM,R3Net} also show its effectiveness in saliency detection. The ASPP uses multiple  parallel  atrous  convolutional layers  with  different  dilation  rates. 
	The sparsity of atrous convolution kernel, especially when using a large dilation rate, results in that the association relationships among sampling points are too weak to extract stable features. 
	In this paper, we apply a simple ``Fold'' operation to effectively relieve this issue. We visualize the folded convolution structure in Fig.~\ref{fig:FoldConv}, which not only further enlarges the receptive field but also extends each valid sampling position from an isolate point to a $2\times2$ connected region.
	
	Let $\mathbf{X}$ represent feature maps with the size of $N \times N \times C$ (C is the channel number). We slide a $2 \times 2$ window on $\mathbf{X}$ in stride $2$ and then conduct atrous convolution with kernel size  $K \times K$ in different dilation rates. Fig.~\ref{fig:FoldConv} shows the computational process when $K = 3$ and dilation rate is $2$. Firstly, we collect  $2 \times 2 \times C$ feature points in each window from  $\mathbf{X}$ and then it is stacked by channel direction, we call this operation "Fold", which is shown in Fig.~\ref{fig:FoldConv}\textcircled{\scriptsize 1}. After the fold operation, we can get new feature maps with the size of $N/2 \times N/2 \times 4C$. A point on the new feature maps corresponds to a $2 \times 2$ area on the original feature maps. Secondly, we adopt an atrous convolution with a kernel size of $3 \times 3$ and dilation rate is $2$. 
	Followed by the reverse process of ``Fold'' which is  called ``Unfold'' operation, the final feature maps are obtained. 
	By using the folded atrous convolution, in the process of information transfer across convolution layers, more contexts are merged and the certain local correlation is also preserved, which provides the fault-tolerance capability for subsequent operations.
	
	As shown in Fig.~\ref{fig:GateNet}, the Fold-ASPP is only equipped on the top of the encoder network, which consists of three folded convolutional layers with  dilation rates $[2, 4, 6]$ to fit the size of feature maps. Just as group convolution~\cite{ResNext} is a trade-off between depthwise convolution~\cite{Xception,Mobilenet} and vanilla convolution in the channel dimension, the proposed folded convolution is a trade-off between atrous convolution and vanilla convolution in the spatial dimension. 
	\subsection{Supervision}\label{sec:supervision}
	As shown in Fig.~\ref{fig:GateNet}, we use the cross-entropy loss for both the intermediate prediction from the FPN branch and the final prediction from the dual branch. In the dual branch decoder, since the FPN branch gradually combines all-level gated encoding and decoding features, it has very powerful prediction ability. We expect that it can predict salient objects as accurately as possible under the supervision of ground truth. While the parallel branch only combines the gated encoding features, which is helpful to remedy the ignored details with the design of residual structure. Moreover, the supervision on $D^{1}$ can drive gate units to learn the weight of the contribution of each encoder block to the final prediction. We use the cross-entropy loss. The total loss \emph{L} could be written as:
	\begin{equation}\label{equ:5}
	\begin{split}
	L = l_{s1}+l_{sf},
	\end{split}
	\end{equation}
	where $l_{s1}$ and $l_{sf}$ are respectively used to regularize the output of the FPN branch and the final prediction. The cross-entropy loss could be computed as:
	\begin{equation}\label{equ:6}
	\begin{split}
	l = YlogP+(1-Y)log(1-P),
	\end{split}
	\end{equation}
	where $P$ and $Y$ denote the predicted map and 
	ground-truth, respectively.
	\section{Experiments}
	\subsection{Experimental Setup}
	\textbf{Dataset.} We evaluate the proposed model
	on five benchmark datasets. \emph{ECSSD}~\cite{ECSSD}  contains $1,000$   semantically meaningful and complex images with pixel-accurate ground truth annotations. \emph{HKU-IS}~\cite{HKU-IS} has $4,447$ challenging  images with multiple disconnected salient objects, overlapping the image boundary. \emph{PASCAL-S}~\cite{PASCAL-S} contains $850$ images selected from the PASCAL VOC 2009 segmentation dataset. \emph{DUT-OMRON}~\cite{DUT-OMRON} includes $5,168$ challenging images, each of which usually has complicated background and one or more foreground objects.
	\emph{DUTS}~\cite{DUTS} is the largest salient object detection dataset, which contains $10,553$ training and $5,019$ test images. These images contain very complex scenarios with high-diversity contents.
	\\
	\textbf{Evaluation Metrics.} For quantitative evaluation, we adopt four widely-used metrics: precision-recall (PR) curve, F-measure score, mean absolute error (MAE) and S-measure score. \emph{Precision-Recall curve:} The pairs of precision and recall are calculated by comparing the binary saliency maps with the ground truth to plot the PR curve, where the threshold for binarizing slides from $0$ to $255$. The closer the PR curve is to the upper right corner, the better the performance is.
	\emph{F-measure}: It is an overall performance measurement that synthetically considers both precision and recall:
	\begin{equation}\label{equ:4}
	\text{F}{_\beta } = \frac{{\left( {1 + {\beta ^2}} \right) \cdot \text{precision} \cdot \text{recall}}}{{{\beta ^2} \cdot \text{precision} + \text{recall}}},
	\end{equation}
	where $\beta^2$ is set to $0.3$ as suggested in~\cite{colorcontrast/Fm} to emphasize the precision. In this paper, we report the maximum F-measure score across the binary maps of different thresholds.
	\emph{Mean Absolute Error}: As the supplement of the PR curve and F-measure, it computes the average absolute difference between the saliency map and the ground truth pixel by pixel.
	\emph{S-measure}: It is more sensitive to foreground structural information than the F-measure. It considers the region-aware structural similarity  $S _ { r }$ and the object-aware structural similarity  $S _ { o }$: 
	\begin{equation}\label{equ:5}
	\text{S}{_m} = \alpha * S _ { o } + ( 1 - \alpha ) * S _ { r },
	\end{equation}
	where $\alpha$ is set to $0.5$~\cite{S-m}.
	\\
	\textbf{Implementation Details.} 
	We follow most state-of-the-art saliency detection methods~\cite{BANet,BASNet,CTBIN,CPD,DGRL,PASE,HRS,PAGRN,BMPM} to use the DUTS-TR as the training dataset which contains $10,553$ images. Our model is implemented based on the Pytorch repository and the hyper-parameters are set as follows: We train the GateNet on a PC with GTX 1080 Ti GPU for $40$ epochs with mini-batch size $4$. For the optimizer, we adopt the stochastic gradient descent (SGD). The momentum, weight decay, and learning rate are set as $0.9$, $0.0005$ and $0.001$, respectively. The ``poly'' policy~\cite{poly} with the power of $0.9$ is used to adjust the learning rate. 
	We adopt some data augmentation techniques to avoid overfitting and make the learned model more robust, which include random horizontally flipping, random rotation, random brightness, saturation and contrast changing. In order to preserve the integrity of the image semantic information, we only resize the image to  $384 \times 384$ instead of using a random crop.
	\subsection{Performance Comparison with State-of-the-art}
	We compare the proposed algorithm with seventeen state-of-the-art saliency detection methods, including the DCL~\cite{DCL}, DSS~\cite{DSS}, Amulet~\cite{Amulet}, SRM~\cite{SRM}, DGRL~\cite{DGRL}, RAS~\cite{RAS}, PAGRN~\cite{PAGRN}, BMPM~\cite{BMPM}, R3Net~\cite{R3Net}, HRS~\cite{HRS}, MLMS~\cite{MLMS}, PAGE~\cite{PASE}, ICNet~\cite{CTBIN}, CPD~\cite{CPD}, BANet~\cite{BANet}, BASNet~\cite{BASNet} and Capsal~\cite{Capsal}. For fair comparisons, all the saliency map of these methods are directly provided by their respective authors or computed by their released codes. To further show the effectiveness of our GateNet, we test its performance in both \textbf{RGBD SOD} and \textbf{Video Object Segmentation} tasks and include the results in appendix.
	
\begin{figure*}
		\centering
		\includegraphics[width=\textwidth]{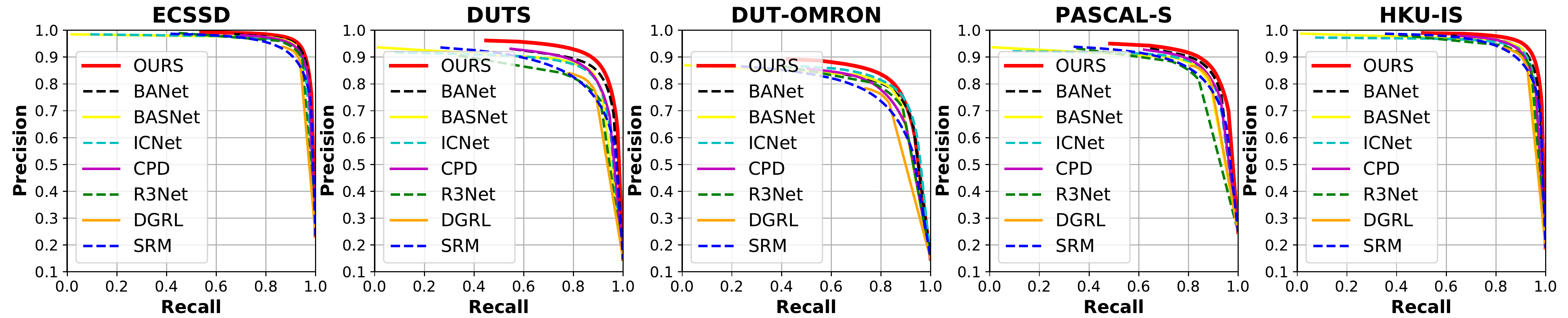}
		\caption{Precision (vertical axis) recall (horizontal axis) curves on six popular rgb-salient object datasets.}
		\label{fig:PR}
	\end{figure*}

	\textbf{Quantitative Evaluation.}
	Tab.~\ref{tab:scores} shows the experimental comparison results in terms of the F-measure, S-measure and MAE scores, from which we can see that the GateNet can consistently outperform other approaches across all five datasets and different metrics.  
	In particular, the GateNet achieves significant performance improvement in terms of the F-measure compared to the second best method BANet~\cite{BANet} on the challenging DUTS-test ($0.870$ vs $0.852$ and $0.888$ vs $0.872$) and PASCAL-S ($0.882$ vs $0.866$ and $0.883$ vs $0.877$) datasets. This clearly demonstrates its superior performance in complex scenes.
	Moreover, some methods~\cite{DCL,DSS,SRM,R3Net} apply the post-processing techniques to refine their saliency maps. Our GateNet still performs better than them without any post-processing.
	We evaluate different algorithms using the standard PR curves in Fig.~\ref{fig:PR}. It can be seen that our PR curves are significantly higher than those of other methods on five datasets. 
	
\begin{figure*}
		\centering
		\includegraphics[width=0.98\textwidth,height=0.28\linewidth]{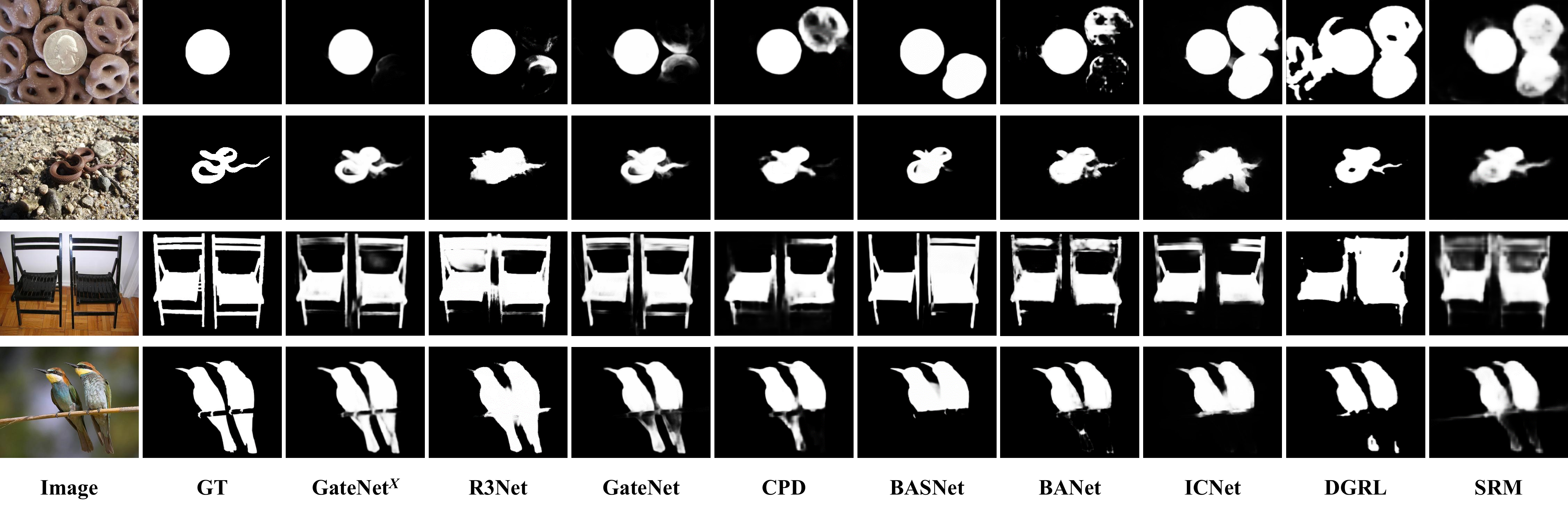}
		\caption{Visual comparison between our results and state-of-the-art methods.}\label{fig:visual-cmp}
	\end{figure*}

	\textbf{Qualitative Evaluation.}
	Fig.~\ref{fig:Qualitative1} and Fig.~\ref{fig:visual-cmp} illustrate some visual comparisons. In Fig.~\ref{fig:Qualitative1}, 
	other methods are severely disturbed by branches and weeds while ours can precisely identify the whole objects. And the GateNet can significantly suppress the background with similar shapes to salient objects (see the $1^{st}$ row in Fig.~\ref{fig:visual-cmp}). Since the Fold-ASPP can obtain more stable structural features, it can help to accurately locate objects and separate adjacent objects well, but some competitors make adjacent objects stick together (see the $3^{th}$ and $4^{th}$ rows in Fig.~\ref{fig:visual-cmp}). Besides,  the proposed parallel branch can restore more details, therefore, the boundary information is retained well.
	
	\subsection{Ablation Studies}
	We detail the contribution of each component to the overall network.
	
	\textbf{Effectiveness of Backbones.} Tab.~\ref{tab:scores} demonstrates that the performance of the gated network can be significantly improved by using better backbones such as ResNet-50, ResNet-101 or ResNeXt-101.
	
	\begin{table*}
		\caption{
			Quantitative comparisons. \textBC{blue}{Blue} indicates the best performance under each backbone setting, while \textBC{red}{red} indicates the best performance among all settings. The subscript in the first column regards the publication year. ``$\dagger$'', ``$S$'' and ``$X$'' mean using the post-processing, ResNet-101 and ResNeXt-101 backbone, respectively. ``---'' represents that the results are not available. $\uparrow$ and $\downarrow$ indicate that the larger and smaller scores are better, respectively.
		}
		\label{tab:scores}
		\renewcommand\tabcolsep{4.0pt} 
		\renewcommand\arraystretch{1.2}
		\scriptsize
		\centering
		\scalebox{.72}
		{
			\begin{tabular}{cccccccccccccccc}
				
				\toprule[2pt]
				\multirow{2}{*}{Method}  &\multicolumn{3}{c}{DUTS-test}&\multicolumn{3}{c}{DUT-OMRON}&\multicolumn{3}{c}{PASCAL-S}&\multicolumn{3}{c}{HKU-IS}&\multicolumn{3}{c}{ECSSD}\\
				\cmidrule(r){2-4} \cmidrule(r){5-7} \cmidrule(r){8-10} \cmidrule(r){11-13} \cmidrule(r){14-16}
				&F$_\beta\uparrow$ & S$_m\uparrow$ & MAE$\downarrow$ & F$_\beta\uparrow$ & S$_m\uparrow$ & MAE$\downarrow$ &F$_\beta\uparrow$ & S$_m\uparrow$ & MAE$\downarrow$ & F$_\beta\uparrow$ & S$_m\uparrow$ & MAE$\downarrow$ &F$_\beta\uparrow$ & S$_m\uparrow$ & MAE$\downarrow$ \\
				\midrule[1pt]
				\multicolumn{16}{c}{VGG-16 backbone} \\
				\midrule[1pt]
				
				DCL$_{16}^\dagger$    & 0.782  & 0.796 & 0.088   & 0.757  & 0.770  & 0.080   & 0.829  & 0.793  & 0.109   & 0.907  & 0.877 & 0.048   & 0.901  & 0.868  & 0.068  \\

				DSS$_{17}^\dagger$     &  ---  &  --- & ---   & 0.781  & 0.789  & 0.063   & 0.840  & 0.792  & 0.098   & 0.916  & 0.878 & 0.040   & 0.921  & 0.882  & 0.052  \\

				Amulet$_{17}$      & 0.778  & 0.804 & 0.085   & 0.743  & 0.780  & 0.098   & 0.839  & 0.819  & 0.099   & 0.899  & 0.886 & 0.050   & 0.915  & 0.894  & 0.059  \\
				
				BMPM$_{18}$ & 0.852 & 0.860 & 0.049 & 0.774 & 0.808 & 0.064 & 0.862 & 0.842  & 0.076 & 0.921 & 0.906 & 
				0.039 
				& 0.928 & 0.911  & 0.045 \\
				RAS$_{18}$     & 0.831  & 0.838 & 0.059   & 0.786  & 0.813  & 0.062  & 0.836 &0.793 &0.106 & 0.913  & 0.887  & 0.045  & 0.921  & 0.893  & 0.056    \\

				PAGRN$_{18}$  & 0.854 & 0.837  & 0.056  & 0.771 & 0.774  & 0.071 & 0.855 & 0.814 & 0.095 & 0.919 & 0.889 & 0.048 & 0.927 & 0.889 & 0.061 \\

				HRS$_{19}$ &  0.843 & 0.828 & 0.051   & 0.762  & 0.771  & 0.066    & 0.850  & 0.798 & 0.092 & 0.913  & 0.882  & 0.042    & 0.920  & 0.883 & 0.054 \\
				MLMS$_{19}$  & 0.852    & 0.861  & 0.049   & 0.774  & 0.808  & 0.064  & 0.864 & 0.844  &0.075   & 0.921  & 0.906 & 0.039   & 0.928 & 0.911  &0.045  \\
				PAGE$_{19}$  & 0.838  & 0.853  & 0.052   & 0.792  &\textBC{blue} {0.824}  & 0.062  & 0.858 & 0.837  & 0.079   & 0.920  & 0.904 & 0.036   & 0.931 & 0.912  & 0.042  \\
				
				BANet$_{19}$ & 0.852 & 0.860 & 0.046 & 0.793 & 0.822&\textBC{blue}{0.061} & 0.866 & 0.838  &0.079 & 0.919 & 0.901 & 0.037 & 0.935 & 0.913  &\textBC{blue}{0.041} \\
				
				GateNet &\textBC{blue}{0.870}& \textBC{blue}{0.869} &\textBC{blue}{0.045} &\textBC{blue}{0.794} &0.820 &\textBC{blue}{0.061} & \textBC{blue}{0.882} &\textBC{blue}{0.855} &\textBC{blue}{0.070}& \textBC{blue}{0.928} & \textBC{blue}{0.909}&\textBC{blue}{0.035} &\textBC{blue}{0.941} & \textBC{blue}{0.917} &\textBC{blue} {0.041}\\
				\midrule[1pt]
				\multicolumn{16}{c}{ResNet-50 backbone} \\
				\midrule[1pt]
				SRM$_{17}^\dagger$  & 0.826 & 0.835  & 0.059  & 0.769 & 0.797  & 0.069 & 0.848 & 0.830 & 0.087 & 0.906 & 0.886 & 0.046 & 0.917 & 0.895 & 0.054 \\
				DGRL$_{18}$  & 0.828 & 0.841  & 0.050  & 0.774 & 0.805  & 0.062 & 0.856 & 0.836 & 0.073 & 0.911 & 0.895 & 0.036 & 0.922 & 0.903 & 0.041 \\
				CPD$_{19}$ & 0.865 & 0.868 & 0.043   & 0.797  & 0.824  & 0.056   & 0.870 & 0.844  & 0.074   & 0.925  & 0.906 & 0.034  & 0.939 & 0.918 &0.037 \\
				ICNet$_{19}$ & 0.855 & 0.864 & 0.048 & 0.813 & \textBC{blue}{0.837} & 0.061 & 0.865 & 0.849  &0.072 & 0.925 & 0.908 & 0.037 & 0.938 & 0.918  & 0.041 \\
				BASNet$_{19}$ & 0.860 & 0.864 & 0.048   & 0.805  & 0.835  & 0.057   & 0.860 & 0.834  & 0.079   & 0.930  & 0.907 & \textBC{blue}{0.033}  & 0.943 & 0.916 &0.037 \\
				BANet$_{19}$ &  0.872 & 0.878 & \textBC{blue}{0.040}   & 0.803  & 0.832  & 0.059   & 0.877  & 0.851  & 0.072   & 0.930  & 0.913 & \textBC{blue}{0.033}  & 0.944  &\textBC{blue}{0.924} &\textBC{blue}{0.035} \\
				
				GateNet &\textBC{blue}{0.888}& \textBC{blue}{0.884} &\textBC{blue}{0.040} &\textBC{blue}{0.818} & \textBC{blue}{0.837} &\textBC{blue}{0.055}& \textBC{blue}{0.883} &\textBC{blue}{0.857} &\textBC{blue}{0.069}& \textBC{blue}{0.933} & \textBC{blue}{0.915}&\textBC{blue}{0.033} &\textBC{blue}{0.945} &0.920&0.040\\
				\midrule[1pt]
				\multicolumn{16}{c}{ResNet/ResNeXt-101 backbone} \\
				\midrule[1pt]
				R3Net$_{18}^{\dagger^X}$  & 0.819 & 0.827 & 0.063   & 0.795  & 0.816  & 0.063   & 0.844 & 0.802  & 0.095   & 0.915  & 0.895 & 0.035  & 0.934 & 0.910 &0.040 \\
				Capsal$_{19}^S$  & 0.819 &  0.818 &  0.063   &  0.639  &  0.673  &  0.101   & 0.869 &  0.837  & 0.074  &  0.883  &  0.851 &  0.058  &  0.863 &  0.826 & 0.077 \\
				GateNet$^S$ & \textBC{blue}{0.893}& \textBC{blue}{0.889} &\textBC{blue}{0.038} &\textBC{blue}{0.821} & \textBC{blue}{0.844} &\textBC{blue}{0.054}& \textBC{blue}{0.883} &\textBC{blue}{0.862} &\textBC{blue}{0.067}& \textBC{blue}{0.937} & \textBC{blue}{0.920}&\textBC{blue}{0.031} &\textBC{blue}{0.951} & \textBC{red}{0.930} &\textBC{red}{0.035}\\
				GateNet$^X$ & \textBC{red}{0.898}& \textBC{red}{0.895} &\textBC{red}{0.035} &\textBC{red}{0.829} & \textBC{red}{0.848} &\textBC{red}{0.051}& \textBC{red}{0.888} &\textBC{red}{0.865} &\textBC{red}{0.065}& \textBC{red}{0.943} & \textBC{red}{0.925}&\textBC{red}{0.029} &\textBC{red}{0.952} & \textBC{blue}{0.929} &\textBC{red}{0.035}\\
				\bottomrule[2pt]
			\end{tabular}
		}
	\end{table*}
	
	\begin{figure*}[ht]
		\centering
		\includegraphics[width=0.60\linewidth]{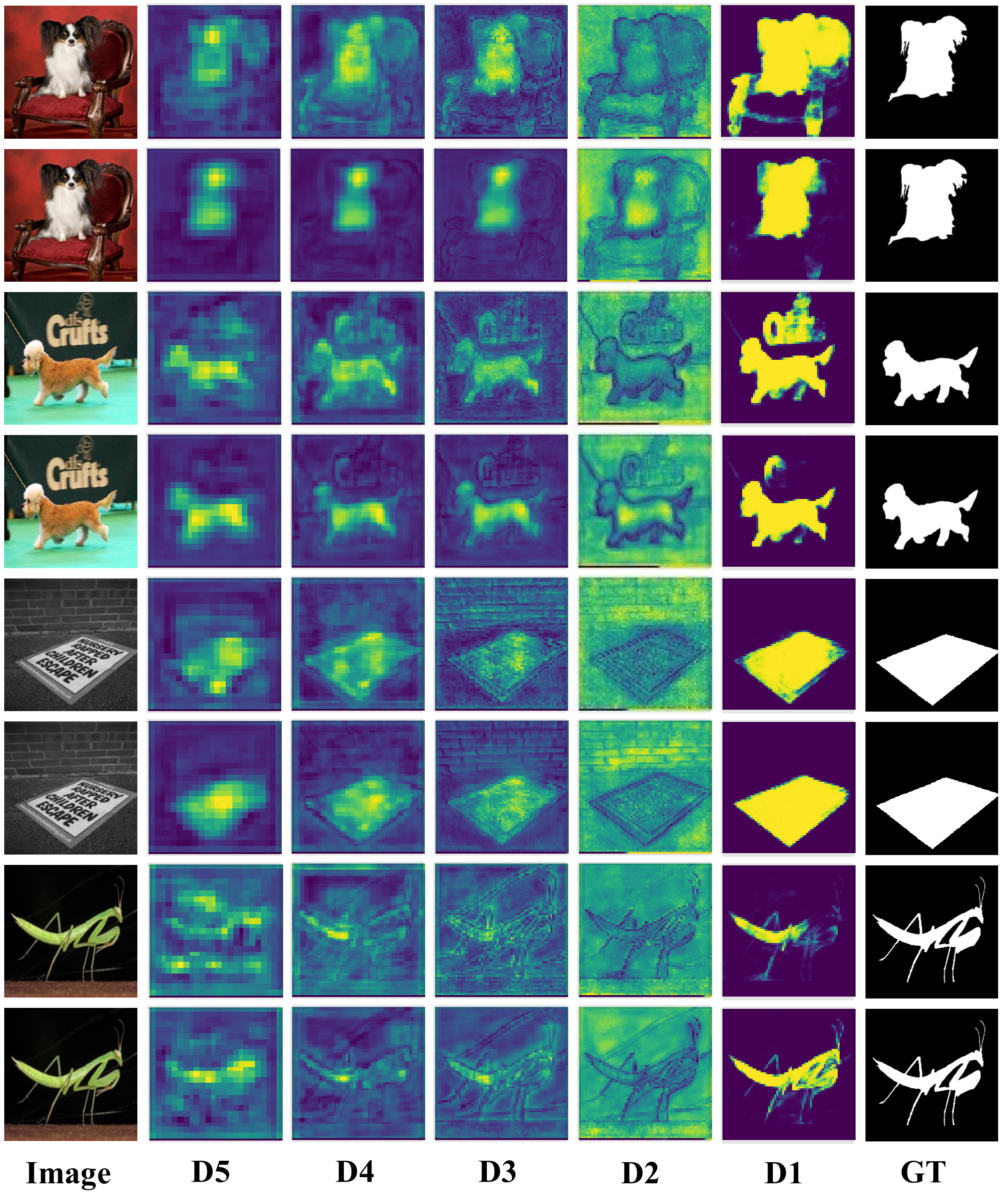}\\
		
		\caption{Visual comparison of feature maps for showing the effect of the multilevel gate units. D5 $\sim$ D1 represent the feature maps of each decoder block from high level to low level. Odd rows and even rows are the results of the FPN baseline without or with multilevel gate units, respectively.}
		\label{fig:Gate_suppress}
	\end{figure*}
	
	\textbf{Effectiveness of Components.} We quantitatively show the benefit of each component in Tab.~\ref{tab:ablation}. We take the results of the VGG-16 backbone with the FPN branch as the baseline. 
	Firstly, the multilevel gate units are added to the baseline network. The performance is significantly improved with the gain of $2.94\%$, $2.17\%$ and $11.67\%$ in terms of the F-measure, S-measure and MAE, respectively. To show the effect of the gate units more intuitively, we visualize the features of different levels in Fig.~\ref{fig:Gate_suppress}.
	It can be observed that even if the dog has a very low contrast with the chair or the billboard (see the $1^{st}$ $\sim$ $4^{th}$ rows), through using multilevel gate units, the high contrast between the object region and the background is always maintained at each layer while the detail information is continually regained, thereby making salient objects be effectively distinguished. 
	Besides, the gate units can avoid excessive suppression for the slender parts of objects (see the $5^{th}$ $\sim$ $8^{th}$ rows). The corners of the poster, the limbs and even tentacles of the mantis are retained well. 
	Secondly, based on the gated baseline network, we design a series of experimental options to verify the effectiveness of the folded convolution and Fold-ASPP. 
	\begin{table*}
		\centering
		\caption{Ablation analysis on the DUTS dataset.} 
		\label{tab:ablation}
		\begin{tabular}{|p{3cm}<{\centering}|p{0.9cm}<{\centering}|p{0.9cm}<{\centering}|p{0.9cm}<{\centering}|} 
			\hline
			
			& F$_\beta$ & S$_m$ & MAE  \\ 
			\hline
			$Baseline$ $(FPN)$&0.816&0.829&0.060 \\
			\hline
			$+$ $Gate$ $Units$ &0.840&0.847&0.053 \\
			\hline
			$+$ $Fold$-$ASPP$ &0.866&0.863&0.047\\
			\hline
			$+$ $Parallel$ $Branch$ &0.870&0.869&0.045\\
			\hline
		\end{tabular}
	\end{table*}
	\begin{table*}
		\caption{Evaluation of the folded convolution and Fold-ASPP. (x) stands for different sampling rates of atrous convolution.}
		\centering
		\label{tab:fold-aspp}
		\scalebox{1}
		{\begin{tabu}{ccccccccc}
				\toprule[2pt]
				& Atrous(2) & Atrous(4) & Atrous(6) & Fold(2) & Fold(4)  & Fold(6) & ASPP  & Fold-ASPP \\
				\midrule[1pt]
				F$_\beta$      & 0.840      & 0.845              & 0.848             & 0.853                & 0.856                 & 0.860      & 0.856          & \textbf{0.866}          \\
				MAE      & 0.055      & 0.053              & 0.051             & 0.051                 & 0.050                 & 0.048      & 0.051   & \textbf{0.047}          \\
				S$_m$    & 0.847      & 0.849              & 0.851             & 0.856                 & 0.858                 & 0.859      & 0.860   & \textbf{0.863}        \\
				\bottomrule[2pt]
			\end{tabu}
		}
	\end{table*}
	Tab.~\ref{tab:fold-aspp} illustrates the results in detail. We adopt the atrous convolution with  dilation rates of $[2, 4, 6]$ and the same dilation rates are also applied to the folded convolution. 
	It can be observed that the folded convolution consistently yields significant performance improvement at each dilation rate than the corresponding atrous convolution in terms of all three metrics. And the single-layer Fold(6) already performs better than the ASPP of aggregating three atrous convolution layers. 
	The Fold-ASPP also naturally outperforms the ASPP with the gain of $1.17\%$ and $8.0\%$ in terms of the F-measure and MAE, respectively.
	Finally, we add the parallel branch to further restore the details of objects. In this process, the gate units, Fold-ASPP and parallel branch complement each other without repulsion.

	\section{Conclusions}
	In this paper, we propose a novel 
	gated network architecture for saliency detection. 
	We first adopt multilevel gate units to balance the contribution of each encoder block and suppress the activation of the features of non-salient regions, which can provide useful context information for the decoder while minimizing interference. The gate unit is simple yet effective, therefore, a gated FPN network can be used as a new baseline for dense prediction tasks. 
	Next, we use the Fold-ASPP to gather multiscale semantic information for the decoder. By the folded operation, the atrous convolution achieves a local-in-local effect, which not only expands the receptive field  
	but also retains the correlation among local sampling points. 
	Finally, to further supplement the details, we combine all encoder features in parallel and construct a residual structure. 
	Experimental results on five benchmark datasets demonstrate that the proposed model outperforms seventeen state-of-the-art methods under different evaluation metrics.
	\\
	\\
	\textbf{Acknowledgements.}
	This work was supported in part by the National Natural Science Foundation of China
	\#61876202, \#61725202, \#61751212 and \#61829102,
	the Dalian Science and Technology Innovation Foundation \#2019J12GX039, and the Fundamental Research Funds for the Central Universities \# DUT20ZD212.
	\appendix
	\section{Appendix}
	 We expand our GateNet to other tasks including RGB-D Salient Object Detection (SOD) and Video Object Segmentation (VOS) to further demonstrate its effectiveness.
	\subsection{Network Architecture}
\begin{figure*}
		\includegraphics[width=\linewidth]{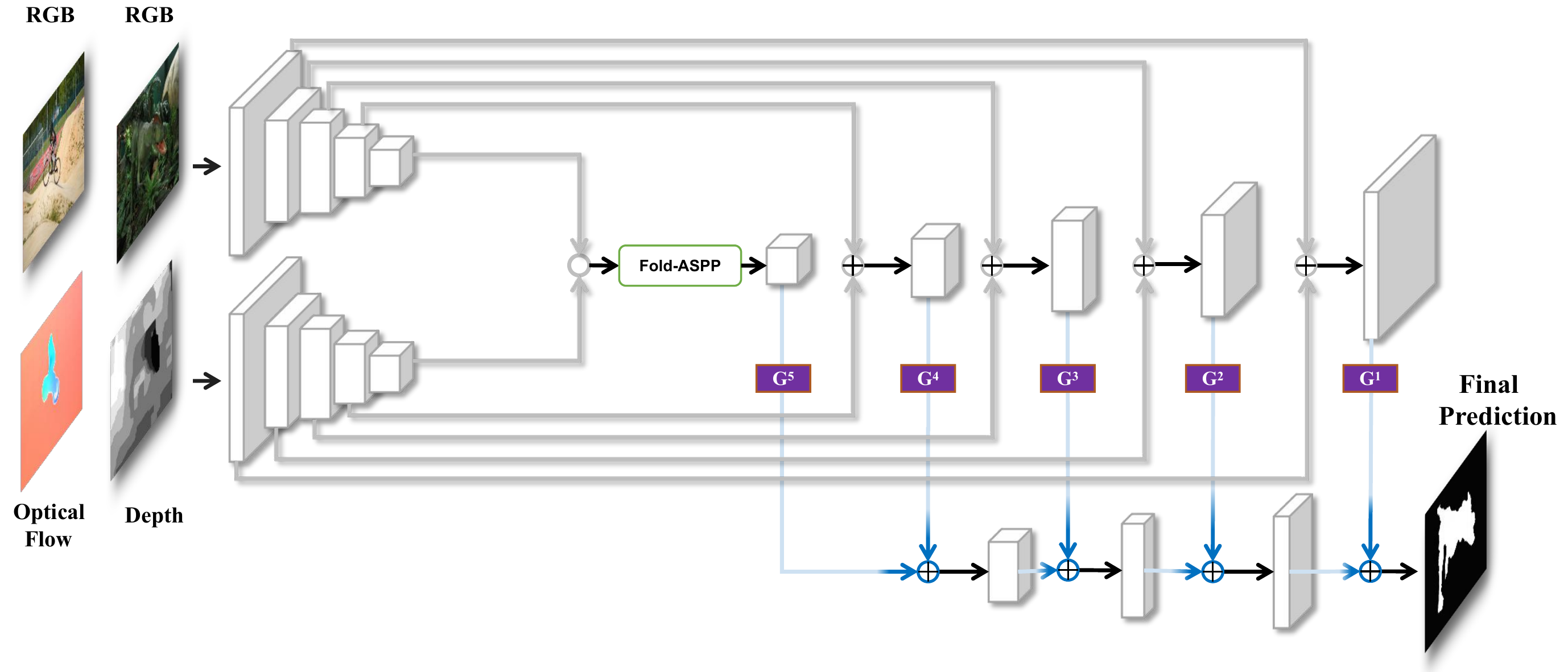}\\ 
		\centering
		\caption{Network pipeline.} 		
		\label{fig:Dual_gatenet}
	\end{figure*}

	Fig.~\ref{fig:Dual_gatenet} shows our proposed dual-branch gated FPN network for RGB-D SOD and VOS. Compared with the RGB SOD network, we only add an extra encoder to extract features of other modals such as depth or optical flow. This dual-branch GateNet is easy to follow and can be used as a new baseline.
	
	\subsection{RGB-D Salient object detection}
	\textbf{Dataset.} There are five main RGB-D SOD datasets which are \textbf{NJUD}~\cite{NJU2000}, \textbf{RGBD135}~\cite{RGBD135} \textbf{NLPR}~\cite{early_fusion_1}, \textbf{SSD}~\cite{SSD} and \textbf{SIP}~\cite{SIP}. 
	We adopt the same splitting way as ~\cite{PCA,MMCI,CTMF,CPFP,DMRA} to guarantee a fair comparison. We split 1,485 samples from NJUD and 700 samples from NLPR for traing a new model. The remaining images in these two datasets and other three datasets are all for testing to verify the generalization ability of saliency models.
	\\
	\textbf{Evaluation Metrics.} We adopt several metrics widely used in RGB-D SOD for quantitative evaluation: F-measure score, mean absolute error (MAE, $\mathcal{M}$), the recently released S-measure ($S_{m}$)~\cite{S-m} and E-measure ($E_{m}$)~\cite{Em} scores. The lower value is better for the MAE and higher is better for others.
	\\
	\textbf{Comparison with State-of-the-art Results.}
	The performance of the proposed model is compared
	with ten state-of-the-art approaches on five benchmark datasets, including the DES~\cite{RGBD135}, DCMC~\cite{DCMC}, CDCP~\cite{CDCP}, DF~\cite{DF}, CTMF~\cite{CTMF}, PCA~\cite{PCA}, MMCI~\cite{MMCI}, TANet~\cite{TANet}, CPFP~\cite{CPFP} and DMRA~\cite{DMRA}. For fair comparisons, all the saliency maps of these methods are directly provided by authors or computed by their released codes. And we take the VGG-16 as the backbone for each stream. 
	Tab.~\ref{tab:rgbd} shows performance comparisons in terms of the maximum F-measure, mean F-measure, weighted F-measure, S-measure, E-measure and MAE scores. It can be seen that our GateNet is very competitive. We believe that future works based on GateNet can further improve performance and easily become the state-of-the-art RGB-D SOD model.
	\subsection{Video Object Segmentation}
	According to whether the mask of the first frame of the video is provided during the test, video object segmentation (vos) can be divided into zero-shot vos and one-shot vos. In this paper, we mainly use the dual-branch GateNet structure as shown in Fig.~\ref{fig:Dual_gatenet} for zero-shot vos.
	\\  
	\textbf{Dataset and Metrics.}
	\textbf{DAVIS-16}~\cite{davis16} is one of the most popular benchmark datasets for video object segmentation tasks. It consists  of  50 high-quality  video  sequences (30 for training and 20 for validation) in total. We follow the training strategy as AGS~\cite{AGS}, COSNet~\cite{COSNet}, PDB~\cite{PDB} and MATNet~\cite{MATNet} to use extra datasets. We use the image saliency datasets: MSRA10K~\cite{MSRA10K} and DUT-OMRON~\cite{DUT-OMRON} to pretrain our RGB branch, then train the whole model with the training videos in DAVIS16. For quantitative evaluation, we adopt two metrics, namely region similarity $\mathcal{J}$ and boundary accuracy $\mathcal{F}$.
	\\
	\textbf{Comparison with State-of-the-art Results.}
	The performance of the proposed model is compared
	with ten state-of-the-art approaches on the DAVIS-16 dataset, including the LVO~\cite{LVO}, ARP~\cite{ARP}, PDB~\cite{PDB}, LSMO~\cite{LSMO}, MotAdapt~\cite{MotAdapt}, EPO~\cite{EPO}, AGS~\cite{AGS}, COSNet~\cite{COSNet}, AnDiff~\cite{AnDiff} and MATNet~\cite{MATNet}. We follow most methods~\cite{MATNet,AnDiff,COSNet,LSMO} to take the ResNet-101 as the backbone. Tab.~\ref{tab:vos} shows performance comparisons in terms of the $\mathcal{J}$ and $\mathcal{F}$. It should be noted that our method only performs feature extraction on the optical flow map generated by PWCNet~\cite{PWC} in order to supplement the motion information of the current frame. Without adding more cross-modal fusion techniques, or using other tracking or detection models, our GateNet can achieve competitive performance with most zero-shot vos methods.
	\begin{table*}[ht]
		\large
		\caption{
			Quantitative comparison. $\uparrow$ and $ \downarrow$ indicate that the larger and smaller scores are better, respectively. Among the CNN-based methods, the best results are shown in $\textBC{red}{red}$. The subscript in each model name is the publication year.
		}
		\label{tab:rgbd}
		\renewcommand\tabcolsep{5.0pt} 
		\renewcommand\arraystretch{1.5}
		\centering
		
		\resizebox{0.92\textwidth}{!}  
		{
			\begin{tabular}{ll|lll|lllllll|l}
				
				\toprule[2pt]
				
				\multicolumn{2}{l|}{\multirow{2}{*}{Metric}} & \multicolumn{3}{c|}{\textbf{\footnotesize{Traditional Methods}}} & \multicolumn{7}{c|}{\textbf{\footnotesize{CNNs-Based Models}}} \\
				\multicolumn{2}{l|}{}   & \Large{DES$_{14}$}         & \Large{DCMC$_{16}$}         &\Large{ CDCP$_{17}$}     &\Large{DF$_{17}$} &\Large{ CTMF$_{18}$} & \Large{PCANet$_{18}$} & \Large{MMCI$_{19}$} &\Large{TANet$_{19}$} & \Large{CPFP$_{19}$} &\Large{ DMRA$_{19}$} &\Large{GateNet}   \\
				\multicolumn{2}{l|}{}   
				&  \multicolumn{1}{c}{\Large{~\cite{RGBD135}}}         &    \multicolumn{1}{c}{\Large{~\cite{DCMC}}}           &   \multicolumn{1}{c|}{\Large{~\cite{CDCP}}}           &  \multicolumn{1}{c}{\Large{~\cite{DF}}}   & \multicolumn{1}{c}{\Large{~\cite{CTMF}}}      &  \multicolumn{1}{c}{\Large{~\cite{PCA}}}       &   \multicolumn{1}{c}{\Large{~\cite{MMCI}}}    &   \multicolumn{1}{c}{\Large{~\cite{TANet}}}     &   \multicolumn{1}{c}{\Large{~\cite{CPFP}}}    &    \multicolumn{1}{c|}{\LARGE{~\cite{DMRA}}}       &  \multicolumn{1}{c}{Ours}  \\
				\hline
				\multirow{6}{*}{\emph{\rotatebox{90}{SSD~\cite{SSD}}}}      
				&$F_{\beta}^{max}\uparrow$ & \multicolumn{1}{c}{\Large{0.260}} &  \multicolumn{1}{c}{\Large{0.750}}    & \multicolumn{1}{c|}{\Large{0.576}}   &  \multicolumn{1}{c}{\Large{0.763}}   &   \multicolumn{1}{c}{\Large{0.755}}    & \multicolumn{1}{c}{\Large{0.844}}  &\multicolumn{1}{c}{\Large{0.823}}  &  \multicolumn{1}{c}{\Large{0.835}}      &  \multicolumn{1}{c}{\Large{0.801}}     &    \multicolumn{1}{c|}{\Large{0.858}}     &   \multicolumn{1}{c}{\textBC{red}{\Large{0.868}}}     \\
				&$F_{\beta}^{mean}\uparrow$ & \multicolumn{1}{c}{\Large{0.073}} &  \multicolumn{1}{c}{\Large{0.684}}    & \multicolumn{1}{c|}{\Large{0.524}}   &  \multicolumn{1}{c}{\Large{0.709}}   &   \multicolumn{1}{c}{\Large{0.709}}    & \multicolumn{1}{c}{\Large{0.786}}  &\multicolumn{1}{c}{\Large{0.748}}  &  \multicolumn{1}{c}{\Large{0.767}}      &  \multicolumn{1}{c}{\Large{0.726}}     &     \multicolumn{1}{c|}{\Large{0.821}}     &   \multicolumn{1}{c}{\textBC{red}{\Large{0.822}}}    \\
				&$F_{\beta}^{w}\uparrow$   & \multicolumn{1}{c}{\Large{0.172}} &  \multicolumn{1}{c}{\Large{0.480}}    & \multicolumn{1}{c|}{\Large{0.429}}   &  \multicolumn{1}{c}{\Large{0.536}}   &   \multicolumn{1}{c}{\Large{0.622}}    & \multicolumn{1}{c}{\Large{0.733}}  &\multicolumn{1}{c}{\Large{0.662}}  &  \multicolumn{1}{c}{\Large{0.727}}      &  \multicolumn{1}{c}{\Large{0.709}}     &     \multicolumn{1}{c|}{\textBC{red}{\Large{0.787}}}     &   \multicolumn{1}{c}{\Large{0.785}}      \\
				& $S_m\uparrow$        & \multicolumn{1}{c}{\Large{0.341}} &  \multicolumn{1}{c}{\Large{0.706}}    & \multicolumn{1}{c|}{\Large{0.603}}   &  \multicolumn{1}{c}{\Large{0.741}}   &   \multicolumn{1}{c}{\Large{0.776}}    & \multicolumn{1}{c}{\Large{0.842}} &\multicolumn{1}{c}{\Large{0.813}}  &  \multicolumn{1}{c}{\Large{0.839}}      &  \multicolumn{1}{c}{\Large{0.807}}     &     \multicolumn{1}{c|}{\Large{0.856}}     &   \multicolumn{1}{c}{\textBC{red}{\Large{0.870}}}   \\
				& $E_m\uparrow$     & \multicolumn{1}{c}{\Large{0.475}} &  \multicolumn{1}{c}{\Large{0.790}}    & \multicolumn{1}{c|}{\Large{0.714}}   &  \multicolumn{1}{c}{\Large{0.801}}   &   \multicolumn{1}{c}{\Large{0.838}}    & \multicolumn{1}{c}{\Large{0.890}}  &\multicolumn{1}{c}{\Large{0.860}}  &  \multicolumn{1}{c}{\Large{0.886}}      &  \multicolumn{1}{c}{\Large{0.832}}     &     \multicolumn{1}{c|}{\Large{0.898}}     &   \multicolumn{1}{c}{\textBC{red}{\Large{0.901}}}    \\
				&$\mathcal{M}\downarrow$ & \multicolumn{1}{c}{\Large{0.500}} &  \multicolumn{1}{c}{\Large{0.168}}    & \multicolumn{1}{c|}{\Large{0.219}}   &  \multicolumn{1}{c}{\Large{0.151}}   &   \multicolumn{1}{c}{\Large{0.100}}    & \multicolumn{1}{c}{\Large{0.063}}  &\multicolumn{1}{c}{\Large{0.082}}  &  \multicolumn{1}{c}{\Large{0.063}}      &  \multicolumn{1}{c}{\Large{0.082}}     &     \multicolumn{1}{c|}{\Large{0.059}}     &   \multicolumn{1}{c}{\textBC{red}{\Large{0.055}}}   \\
				
				\hline
				\multirow{6}{*}{\emph{\rotatebox{90}{NJUD~\cite{NJU2000}}}}      
				&$F_{\beta}^{max}\uparrow$  & \multicolumn{1}{c}{\Large{0.328}} &  \multicolumn{1}{c}{\Large{0.769}}    & \multicolumn{1}{c|}{\Large{0.661}}   &  \multicolumn{1}{c}{\Large{0.789}}   &   \multicolumn{1}{c}{\Large{0.857}}    & \multicolumn{1}{c}{\Large{0.888}}  &\multicolumn{1}{c}{\Large{0.868}}  &  \multicolumn{1}{c}{\Large{0.888}}      &  \multicolumn{1}{c}{\Large{0.890}}     &     \multicolumn{1}{c|}{\Large{0.896}}     &   \multicolumn{1}{c}{\textBC{red}{\Large{0.914}}}    \\
				&$F_{\beta}^{mean}\uparrow$  & \multicolumn{1}{c}{\Large{0.165}} &  \multicolumn{1}{c}{\Large{0.715}}    & \multicolumn{1}{c|}{\Large{0.618}}   &  \multicolumn{1}{c}{\Large{0.744}}   &   \multicolumn{1}{c}{\Large{0.788}}    & \multicolumn{1}{c}{\Large{0.844}}  &\multicolumn{1}{c}{\Large{0.813}}  &  \multicolumn{1}{c}{\Large{0.844}}      &  \multicolumn{1}{c}{\Large{0.837}}     &     \multicolumn{1}{c|}{\Large{0.871}}     &   \multicolumn{1}{c}{\textBC{red}{\Large{0.879}}}      \\
				&$F_{\beta}^{w}\uparrow$   & \multicolumn{1}{c}{\Large{0.234}} &  \multicolumn{1}{c}{\Large{0.497}}    & \multicolumn{1}{c|}{\Large{0.510}}   &  \multicolumn{1}{c}{\Large{0.545}}   &   \multicolumn{1}{c}{\Large{0.720}}    & \multicolumn{1}{c}{\Large{0.803}}  &\multicolumn{1}{c}{\Large{0.739}}  &  \multicolumn{1}{c}{\Large{0.805}}      &  \multicolumn{1}{c}{\Large{0.828}}     &     \multicolumn{1}{c|}{\Large{0.847}}     &   \multicolumn{1}{c}{\textBC{red}{\Large{0.849}}}       \\
				& $S_m\uparrow$        & \multicolumn{1}{c}{\Large{0.413}} &  \multicolumn{1}{c}{\Large{0.703}}    & \multicolumn{1}{c|}{\Large{0.672}}   &  \multicolumn{1}{c}{\Large{0.735}}   &   \multicolumn{1}{c}{\Large{0.849}}    & \multicolumn{1}{c}{\Large{0.877}} &\multicolumn{1}{c}{\Large{0.859}}  &  \multicolumn{1}{c}{\Large{0.878}}      &  \multicolumn{1}{c}{\Large{0.878}}     &     \multicolumn{1}{c|}{\Large{0.885}}     &   \multicolumn{1}{c}{\textBC{red}{\Large{0.902}}}       \\
				& $E_m\uparrow$     & \multicolumn{1}{c}{\Large{0.491}} &  \multicolumn{1}{c}{\Large{0.796}}    & \multicolumn{1}{c|}{\Large{0.751}}   &  \multicolumn{1}{c}{\Large{0.818}}   &   \multicolumn{1}{c}{\Large{0.866}}    & \multicolumn{1}{c}{\Large{0.909}}  &\multicolumn{1}{c}{\Large{0.882}}  &  \multicolumn{1}{c}{\Large{0.909}}      &  \multicolumn{1}{c}{\Large{0.900}}     &     \multicolumn{1}{c|}{\Large{0.920}}     &   \multicolumn{1}{c}{\textBC{red}{\Large{0.922}}}      \\
				&$\mathcal{M}\downarrow$ & \multicolumn{1}{c}{\Large{0.448}} &  \multicolumn{1}{c}{\Large{0.167}}    & \multicolumn{1}{c|}{\Large{0.182}}   &  \multicolumn{1}{c}{\Large{0.151}}   &   \multicolumn{1}{c}{\Large{0.085}}    & \multicolumn{1}{c}{\Large{0.059}}  &\multicolumn{1}{c}{\Large{0.079}}  &  \multicolumn{1}{c}{\Large{0.061}}      &  \multicolumn{1}{c}{\Large{0.053}}     &     \multicolumn{1}{c|}{\Large{0.051}}     &   \multicolumn{1}{c}{\textBC{red}{\Large{0.047}}}      \\
				\hline
				\multirow{6}{*}{\emph{\rotatebox{90}{RGBD135~\cite{RGBD135}}}}      
				&$F_{\beta}^{max}\uparrow$   & \multicolumn{1}{c}{\Large{0.800}} &  \multicolumn{1}{c}{\Large{0.311}}    & \multicolumn{1}{c|}{\Large{0.651}}   &  \multicolumn{1}{c}{\Large{0.625}}   &   \multicolumn{1}{c}{\Large{0.865}}    & \multicolumn{1}{c}{\Large{0.842}}  &\multicolumn{1}{c}{\Large{0.839}}  &  \multicolumn{1}{c}{\Large{0.853}}      &  \multicolumn{1}{c}{\Large{0.882}}     &     \multicolumn{1}{c|}{\Large{0.906}}     &   \multicolumn{1}{c}{\textBC{red}{\Large{0.919}}}      \\
				&$F_{\beta}^{mean}\uparrow$  & \multicolumn{1}{c}{\Large{0.695}} &  \multicolumn{1}{c}{\Large{0.234}}    & \multicolumn{1}{c|}{\Large{0.594}}   &  \multicolumn{1}{c}{\Large{0.573}}   &   \multicolumn{1}{c}{\Large{0.778}}    & \multicolumn{1}{c}{\Large{0.774}}  &\multicolumn{1}{c}{\Large{0.762}}  &  \multicolumn{1}{c}{\Large{0.795}}      &  \multicolumn{1}{c}{\Large{0.829}}     &     \multicolumn{1}{c|}{\Large{0.867}}     &   \multicolumn{1}{c}{\textBC{red}{\Large{0.891}}}      \\
				&$F_{\beta}^{w}\uparrow$   & \multicolumn{1}{c}{\Large{0.301}} &  \multicolumn{1}{c}{\Large{0.169}}    & \multicolumn{1}{c|}{\Large{0.478}}   &  \multicolumn{1}{c}{\Large{0.392}}   &   \multicolumn{1}{c}{\Large{0.687}}    & \multicolumn{1}{c}{\Large{0.711}}  &\multicolumn{1}{c}{\Large{0.650}}  &  \multicolumn{1}{c}{\Large{0.740}}      &  \multicolumn{1}{c}{\Large{0.787}}     &     \multicolumn{1}{c|}{\textBC{red}{\Large{0.843}}}     &   \multicolumn{1}{c}{\Large{0.838}}       \\
				& $S_m\uparrow$        & \multicolumn{1}{c}{\Large{0.632}} &  \multicolumn{1}{c}{\Large{0.469}}    & \multicolumn{1}{c|}{\Large{0.709}}   &  \multicolumn{1}{c}{\Large{0.685}}   &   \multicolumn{1}{c}{\Large{0.863}}    & \multicolumn{1}{c}{\Large{0.843}} &\multicolumn{1}{c}{\Large{0.848}}  &  \multicolumn{1}{c}{\Large{0.858}}      &  \multicolumn{1}{c}{\Large{0.872}}     &     \multicolumn{1}{c|}{\Large{0.899}}     &   \multicolumn{1}{c}{\textBC{red}{\Large{0.905}}}        \\
				& $E_m\uparrow$     & \multicolumn{1}{c}{\Large{0.817}} &  \multicolumn{1}{c}{\Large{0.676}}    & \multicolumn{1}{c|}{\Large{0.810}}   &  \multicolumn{1}{c}{\Large{0.806}}   &   \multicolumn{1}{c}{\Large{0.911}}    & \multicolumn{1}{c}{\Large{0.912}}  &\multicolumn{1}{c}{\Large{0.904}}  &  \multicolumn{1}{c}{\Large{0.919}}      &  \multicolumn{1}{c}{\Large{0.927}}     &     \multicolumn{1}{c|}{\Large{0.944}}     &   \multicolumn{1}{c}{\textBC{red}{\Large{0.966}}}       \\
				&$\mathcal{M}\downarrow$ & \multicolumn{1}{c}{\Large{0.289}} &  \multicolumn{1}{c}{\Large{0.196}}    & \multicolumn{1}{c|}{\Large{0.120}}   &  \multicolumn{1}{c}{\Large{0.131}}   &   \multicolumn{1}{c}{\Large{0.055}}    & \multicolumn{1}{c}{\Large{0.050}}  &\multicolumn{1}{c}{\Large{0.065}}  &  \multicolumn{1}{c}{\Large{0.046}}      &  \multicolumn{1}{c}{\Large{0.038}}     &     \multicolumn{1}{c|}{\textBC{red}{\Large{0.030}}}     &   \multicolumn{1}{c}{\textBC{red}{\Large{0.030}}}       \\
				
				\hline
				\multirow{6}{*}{\emph{\rotatebox{90}{NLPR~\cite{early_fusion_1}}}}      
				&$F_{\beta}^{max}\uparrow$   & \multicolumn{1}{c}{\Large{0.695}} &  \multicolumn{1}{c}{\Large{0.413}}    & \multicolumn{1}{c|}{\Large{0.687}}   &  \multicolumn{1}{c}{\Large{0.752}}   &   \multicolumn{1}{c}{\Large{0.841}}    & \multicolumn{1}{c}{\Large{0.864}}  &\multicolumn{1}{c}{\Large{0.841}}  &  \multicolumn{1}{c}{\Large{0.876}}      &  \multicolumn{1}{c}{\Large{0.884}}     &     \multicolumn{1}{c|}{\Large{0.888}}     &   \multicolumn{1}{c}{\textBC{red}{\Large{0.904}}}      \\
				&$F_{\beta}^{mean}\uparrow$   & \multicolumn{1}{c}{\Large{0.583}} &  \multicolumn{1}{c}{\Large{0.328}}    & \multicolumn{1}{c|}{\Large{0.592}}   &  \multicolumn{1}{c}{\Large{0.683}}   &   \multicolumn{1}{c}{\Large{0.724}}    & \multicolumn{1}{c}{\Large{0.795}}  &\multicolumn{1}{c}{\Large{0.730}}  &  \multicolumn{1}{c}{\Large{0.796}}      &  \multicolumn{1}{c}{\Large{0.818}}     &     \multicolumn{1}{c|}{\textBC{red}{\Large{0.855}}}     &   \multicolumn{1}{c}{\Large{0.854}}     \\
				&$F_{\beta}^{w}\uparrow$   & \multicolumn{1}{c}{\Large{0.254}} &  \multicolumn{1}{c}{\Large{0.259}}    & \multicolumn{1}{c|}{\Large{0.501}}   &  \multicolumn{1}{c}{\Large{0.516}}   &   \multicolumn{1}{c}{\Large{0.679}}    & \multicolumn{1}{c}{\Large{0.762}}  &\multicolumn{1}{c}{\Large{0.676}}  &  \multicolumn{1}{c}{\Large{0.780}}      &  \multicolumn{1}{c}{\Large{0.807}}     &     \multicolumn{1}{c|}{\textBC{red}{\Large{0.840}}}     &   \multicolumn{1}{c}{\Large{0.838}}       \\
				& $S_m\uparrow  $        & \multicolumn{1}{c}{\Large{0.582}} &  \multicolumn{1}{c}{\Large{0.550}}    & \multicolumn{1}{c|}{\Large{0.724}}   &  \multicolumn{1}{c}{\Large{0.769}}   &   \multicolumn{1}{c}{\Large{0.860}}    & \multicolumn{1}{c}{\Large{0.874}}  &\multicolumn{1}{c}{\Large{0.856}}  &  \multicolumn{1}{c}{\Large{0.886}}      &  \multicolumn{1}{c}{\Large{0.884}}     &     \multicolumn{1}{c|}{\Large{0.898}}     &   \multicolumn{1}{c}{\textBC{red}{\Large{0.910}}}      \\
				& $E_m\uparrow$         & \multicolumn{1}{c}{\Large{0.760}} &  \multicolumn{1}{c}{\Large{0.685}}    & \multicolumn{1}{c|}{\Large{0.786}}   &  \multicolumn{1}{c}{\Large{0.840}}   &   \multicolumn{1}{c}{\Large{0.869}}    & \multicolumn{1}{c}{\Large{0.916}}  &\multicolumn{1}{c}{\Large{0.872}}  &  \multicolumn{1}{c}{\Large{0.916}}      &  \multicolumn{1}{c}{\Large{0.920}}     &     \multicolumn{1}{c|}{\textBC{red}{\Large{0.942}}}     &   \multicolumn{1}{c}{\textBC{red}{\Large{0.942}}}       \\
				&$\mathcal{M}\downarrow$  & \multicolumn{1}{c}{\Large{0.301}} &  \multicolumn{1}{c}{\Large{0.196}}    & \multicolumn{1}{c|}{\Large{0.115}}   &  \multicolumn{1}{c}{\Large{0.100}}   &   \multicolumn{1}{c}{\Large{0.056}}    & \multicolumn{1}{c}{\Large{0.044}}  &\multicolumn{1}{c}{\Large{0.059}}  &  \multicolumn{1}{c}{\Large{0.041}}      &  \multicolumn{1}{c}{\Large{0.038}}     &     \multicolumn{1}{c|}{\textBC{red}{\Large{0.031}}}     &   \multicolumn{1}{c}{\Large{0.032}}      \\
				\hline
				\multirow{6}{*}{\emph{\rotatebox{90}{SIP~\cite{SIP}}}}      
				&$F_{\beta}^{max}\uparrow$  & \multicolumn{1}{c}{\Large{0.720}} &  \multicolumn{1}{c}{\Large{0.680}}    & \multicolumn{1}{c|}{\Large{0.544}}   &  \multicolumn{1}{c}{\Large{0.704}}   &   \multicolumn{1}{c}{\Large{0.720}}    & \multicolumn{1}{c}{\Large{0.861}}  &\multicolumn{1}{c}{\Large{0.840}}  &  \multicolumn{1}{c}{\Large{0.851}}      &  \multicolumn{1}{c}{\Large{0.870}}     &     \multicolumn{1}{c|}{\Large{0.847}}     &   \multicolumn{1}{c}{\textBC{red}{\Large{0.894}}}     \\
				&$F_{\beta}^{mean}\uparrow$  & \multicolumn{1}{c}{\Large{0.644}} &  \multicolumn{1}{c}{\Large{0.645}}    & \multicolumn{1}{c|}{\Large{0.495}}   &  \multicolumn{1}{c}{\Large{0.673}}   &   \multicolumn{1}{c}{\Large{0.684}}    & \multicolumn{1}{c}{\Large{0.825}}  &\multicolumn{1}{c}{\Large{0.795}}  &  \multicolumn{1}{c}{\Large{0.809}}      &  \multicolumn{1}{c}{\Large{0.819}}     &     \multicolumn{1}{c|}{\Large{0.815}}     &   \multicolumn{1}{c}{\textBC{red}{\Large{0.856}}}      \\
				&$F_{\beta}^{w}\uparrow$   & \multicolumn{1}{c}{\Large{0.342}} &  \multicolumn{1}{c}{\Large{0.414}}    & \multicolumn{1}{c|}{\Large{0.397}}   &  \multicolumn{1}{c}{\Large{0.406}}   &   \multicolumn{1}{c}{\Large{0.535}}    & \multicolumn{1}{c}{\Large{0.768}}  &\multicolumn{1}{c}{\Large{0.712}}  &  \multicolumn{1}{c}{\Large{0.748}}      &  \multicolumn{1}{c}{\Large{0.788}}     &     \multicolumn{1}{c|}{\Large{0.734}}     &   \multicolumn{1}{c}{\textBC{red}{\Large{0.810}}}      \\
				& $S_m\uparrow$        & \multicolumn{1}{c}{\Large{0.616}} &  \multicolumn{1}{c}{\Large{0.683}}    & \multicolumn{1}{c|}{\Large{0.595}}   &  \multicolumn{1}{c}{\Large{0.653}}   &   \multicolumn{1}{c}{\Large{0.716}}    & \multicolumn{1}{c}{\Large{0.842}} &\multicolumn{1}{c}{\Large{0.833}}  &  \multicolumn{1}{c}{\Large{0.835}}      &  \multicolumn{1}{c}{\Large{0.850}}     &     \multicolumn{1}{c|}{\Large{0.800}}     &   \multicolumn{1}{c}{\textBC{red}{\Large{0.874}}}       \\
				& $E_m\uparrow$     & \multicolumn{1}{c}{\Large{0.751}} &  \multicolumn{1}{c}{\Large{0.787}}    & \multicolumn{1}{c|}{\Large{0.722}}   &  \multicolumn{1}{c}{\Large{0.794}}   &   \multicolumn{1}{c}{\Large{0.824}}    & \multicolumn{1}{c}{\Large{0.900}}  &\multicolumn{1}{c}{\Large{0.886}}  &  \multicolumn{1}{c}{\Large{0.894}}      &  \multicolumn{1}{c}{\Large{0.899}}     &      \multicolumn{1}{c|}{\Large{0.858}}     &   \multicolumn{1}{c}{\textBC{red}{\Large{0.914}}}       \\
				&$\mathcal{M}\downarrow$ & \multicolumn{1}{c}{\Large{0.298}} &  \multicolumn{1}{c}{\Large{0.186}}    & \multicolumn{1}{c|}{\Large{0.224}}   &  \multicolumn{1}{c}{\Large{0.185}}   &   \multicolumn{1}{c}{\Large{0.139}}    & \multicolumn{1}{c}{\Large{0.071}}  &\multicolumn{1}{c}{\Large{0.086}}  &  \multicolumn{1}{c}{\Large{0.075}}      &  \multicolumn{1}{c}{\Large{0.064}}     &     \multicolumn{1}{c|}{\Large{0.088}}     &   \multicolumn{1}{c}{\textBC{red}{\Large{0.057}}}      \\
				\bottomrule[2pt]
			\end{tabular}
		}
	\end{table*}
	
	\begin{table*}[ht]
		\large
		\caption{
			Quantitative comparison of Zero-shot VOS methods on the DAVIS-16 validation set. $\uparrow$ and $ \downarrow$ indicate that the larger and smaller scores are better, respectively. The best results are shown in $\textBC{red}{red}$. The subscript in each model name is the publication year.
		}
		\label{tab:vos}
		\renewcommand\tabcolsep{5.0pt} 
		\renewcommand\arraystretch{1.5}
		\centering
		
		\resizebox{0.92\textwidth}{!}  
		{
			\begin{tabular}{ll|llllllllll|l}
				
				\toprule[2pt]
				\multicolumn{2}{l|}{\multirow{2}{*}{Metric}}   & \Large{LVO$_{17}$}         & \Large{ARP$_{17}$}         &\Large{PDB$_{18}$}     &\Large{LSMO$_{19}$} &\Large{MotAdapt$_{19}$} & \Large{EPO$_{20}$} & \Large{AGS$_{19}$} &\Large{COSNet$_{19}$} & \Large{AnDiff$_{19}$} &\Large{MATNet$_{20}$} &\Large{GateNet}   \\
				\multicolumn{2}{l|}{}   
				&  \multicolumn{1}{c}{\Large{~\cite{LVO}}}         &    \multicolumn{1}{c}{\Large{~\cite{ARP}}}           &   \multicolumn{1}{c}{\Large{~\cite{PDB}}}           &  \multicolumn{1}{c}{\Large{~\cite{LSMO}}}   & \multicolumn{1}{c}{\Large{~\cite{MotAdapt}}}      &  \multicolumn{1}{c}{\Large{~\cite{EPO}}}       &   \multicolumn{1}{c}{\Large{~\cite{AGS}}}    &   \multicolumn{1}{c}{\Large{~\cite{COSNet}}}     &   \multicolumn{1}{c}{\Large{~\cite{AnDiff}}}    &    \multicolumn{1}{c|}{\LARGE{~\cite{MATNet}}}       &  \multicolumn{1}{c}{Ours}  \\
				\hline
				
				\multirow{3}{*}{$\mathcal{J}$}      
				&Mean$\uparrow$& \multicolumn{1}{c}{\Large{75.9}} &  \multicolumn{1}{c}{\Large{76.2}}    & \multicolumn{1}{c}{\Large{77.2}}   &  \multicolumn{1}{c}{\Large{78.2}}   &   \multicolumn{1}{c}{\Large{77.2}}    & \multicolumn{1}{c}{\Large{80.6}}  &\multicolumn{1}{c}{\Large{79.7}}  &  \multicolumn{1}{c}{\Large{80.5}}      &  \multicolumn{1}{c}{\Large{81.7}}     &    \multicolumn{1}{c|}{\textBC{red}{\Large{82.4}}}     &   \multicolumn{1}{c}{\Large{80.9}}     \\
				&Recall$\uparrow$ & \multicolumn{1}{c}{\Large{89.1}} &  \multicolumn{1}{c}{\Large{91.1}}    & \multicolumn{1}{c}{\Large{90.1}}   &  \multicolumn{1}{c}{\Large{89.1}}   &   \multicolumn{1}{c}{\Large{87.8}}    & \multicolumn{1}{c}{\textBC{red}{\Large{95.2}}}  &\multicolumn{1}{c}{\Large{91.1}}  &  \multicolumn{1}{c}{\Large{93.1}}      &  \multicolumn{1}{c}{\Large{90.9}}     &     \multicolumn{1}{c|}{\Large{94.5}}     &   \multicolumn{1}{c}{\Large{94.3}}    \\
				&Decay$\downarrow$ & \multicolumn{1}{c}{\textBC{red}{\Large{0.0}}} &  \multicolumn{1}{c}{\Large{7.0}}    & \multicolumn{1}{c}{\Large{0.9}}   &  \multicolumn{1}{c}{\Large{4.1}}   &   \multicolumn{1}{c}{\Large{5.0}}    & \multicolumn{1}{c}{\Large{2.2}}  &\multicolumn{1}{c}{\Large{1.9}}  &  \multicolumn{1}{c}{\Large{4.4}}      &  \multicolumn{1}{c}{\Large{2.2}}     &     \multicolumn{1}{c|}{\Large{5.5}}     &   \multicolumn{1}{c}{\Large{3.3}}   \\
				
				\hline
				\multirow{3}{*}{$\mathcal{F}$}      
				&Mean$\uparrow$& \multicolumn{1}{c}{\Large{72.1}} &  \multicolumn{1}{c}{\Large{70.6}}    & \multicolumn{1}{c}{\Large{74.5}}   &  \multicolumn{1}{c}{\Large{75.9}}   &   \multicolumn{1}{c}{\Large{77.4}}    & \multicolumn{1}{c}{\Large{75.5}}  &\multicolumn{1}{c}{\Large{77.4}}  &  \multicolumn{1}{c}{\Large{79.5}}      &  \multicolumn{1}{c}{\Large{80.5}}     &    \multicolumn{1}{c|}{\textBC{red}{\Large{80.7}}}     &   \multicolumn{1}{c}{\Large{79.4}}     \\
				&Recall$\uparrow$ & \multicolumn{1}{c}{\Large{83.4}} &  \multicolumn{1}{c}{\Large{83.5}}    & \multicolumn{1}{c}{\Large{84.4}}   &  \multicolumn{1}{c}{\Large{84.7}}   &   \multicolumn{1}{c}{\Large{84.4}}    & \multicolumn{1}{c}{\Large{87.9}}  &\multicolumn{1}{c}{\Large{85.8}}  &  \multicolumn{1}{c}{\Large{89.5}}      &  \multicolumn{1}{c}{\Large{85.1}}     &     \multicolumn{1}{c|}{\textBC{red}{\Large{90.2}}}     &   \multicolumn{1}{c}{\Large{89.2}}    \\
				&Decay$\downarrow$ & \multicolumn{1}{c}{\Large{1.3}} &  \multicolumn{1}{c}{\Large{7.9}}    & \multicolumn{1}{c}{\textBC{red}{\Large{-0.2}}}   &  \multicolumn{1}{c}{\Large{3.5}}   &   \multicolumn{1}{c}{\Large{3.3}}    & \multicolumn{1}{c}{\Large{2.4}}  &\multicolumn{1}{c}{\Large{1.6}}  &  \multicolumn{1}{c}{\Large{5.0}}      &  \multicolumn{1}{c}{\Large{0.6}}     &     \multicolumn{1}{c|}{\Large{4.5}}     &   \multicolumn{1}{c}{\Large{2.9}}  \\
				\bottomrule[2pt]
			\end{tabular}
		}
	\end{table*}

	\clearpage
	%
	%
	\bibliographystyle{splncs04}
	\bibliography{egbib}
\end{document}